\documentclass[twoside]{article}

\usepackage{PRIMEarxiv}

\usepackage[utf8]{inputenc} 
\usepackage[T1]{fontenc}    
\usepackage{hyperref}       
\usepackage{url}            
\usepackage{booktabs}       
\usepackage{amsfonts}       
\usepackage{nicefrac}       
\usepackage{microtype}      
\usepackage{lipsum}
\usepackage{fancyhdr}       
\usepackage{graphicx}       
\graphicspath{{media/}}     
\usepackage{subcaption}

\usepackage{comment}

\usepackage{algorithm}
\usepackage{algpseudocode}
\usepackage{caption}
\usepackage{nicefrac}
\usepackage{amsmath}
\usepackage{amssymb}
\usepackage{makecell}
\usepackage{caption}

\usepackage[most]{tcolorbox}
\usepackage{listings}
\tcbuselibrary{listings,breakable}

\lstdefinestyle{promptstyle}{
  basicstyle=\ttfamily\small,
  breaklines=true,
  breakatwhitespace=false,
  columns=fullflexible,
  keepspaces=true,
  showstringspaces=false,
}

\newtcblisting{CodeBox}{
  listing only,
  listing options={style=promptstyle},
  breakable,
  enhanced,
  colback=gray!3,
  colframe=gray!60,
  boxrule=0.4pt,
  arc=2mm,
  left=1ex, right=1ex, top=0.8ex, bottom=0.8ex,
  title={},
}

\title{Solving a Million-Step LLM Task with Zero Errors}

\author{
  Elliot Meyerson\thanks{Correspondence to: \texttt{elliot.meyerson@cognizant.com}} \\
  Cognizant AI Lab
  \And
  Giuseppe Paolo \\
  Cognizant AI Lab
  \And
  Roberto Dailey \\
  Cognizant AI Lab
  \And
  Hormoz Shahrzad\\
  UT Austin \& Cognizant AI Lab
  \And
  Olivier Francon\\
  Cognizant AI Lab
  \And
  Conor F. Hayes\\
  Cognizant AI Lab
  \And
  Xin Qiu \\
  Cognizant AI Lab
  \And
  Babak Hodjat \\
  Cognizant AI Lab
  \And
  Risto Miikkulainen \\
  UT Austin \& Cognizant AI Lab
}

\begin{document}
\maketitle

\begin{abstract}
LLMs have achieved remarkable breakthroughs in reasoning, insights, and tool use, but chaining these abilities into extended processes at the scale of those routinely executed by humans, organizations, and societies has remained out of reach.
The models have a persistent error rate that prevents scale-up: for instance, recent experiments in the Towers of Hanoi benchmark domain showed that the process inevitably becomes derailed after at most a few hundred steps.
Thus, although LLM research is often still benchmarked on tasks with relatively few dependent logical steps, there is increasing attention on the ability (or inability) of LLMs to perform long range tasks.
This paper describes MAKER, the first system that successfully solves a task with over one million LLM steps with zero errors, and, in principle, scales far beyond this level.
The approach relies on an extreme decomposition of a task into subtasks, each of which can be tackled by focused microagents.
The high level of modularity resulting from the decomposition allows error correction to be applied at each step through an efficient multi-agent voting scheme. This combination of extreme decomposition and error correction makes scaling possible. Thus, the results suggest that instead of relying on continual improvement of current LLMs, massively decomposed agentic processes (MDAPs) may provide a way to efficiently solve problems at the level of organizations and societies.
\end{abstract}


\begin{center}
    \includegraphics[width=\linewidth]{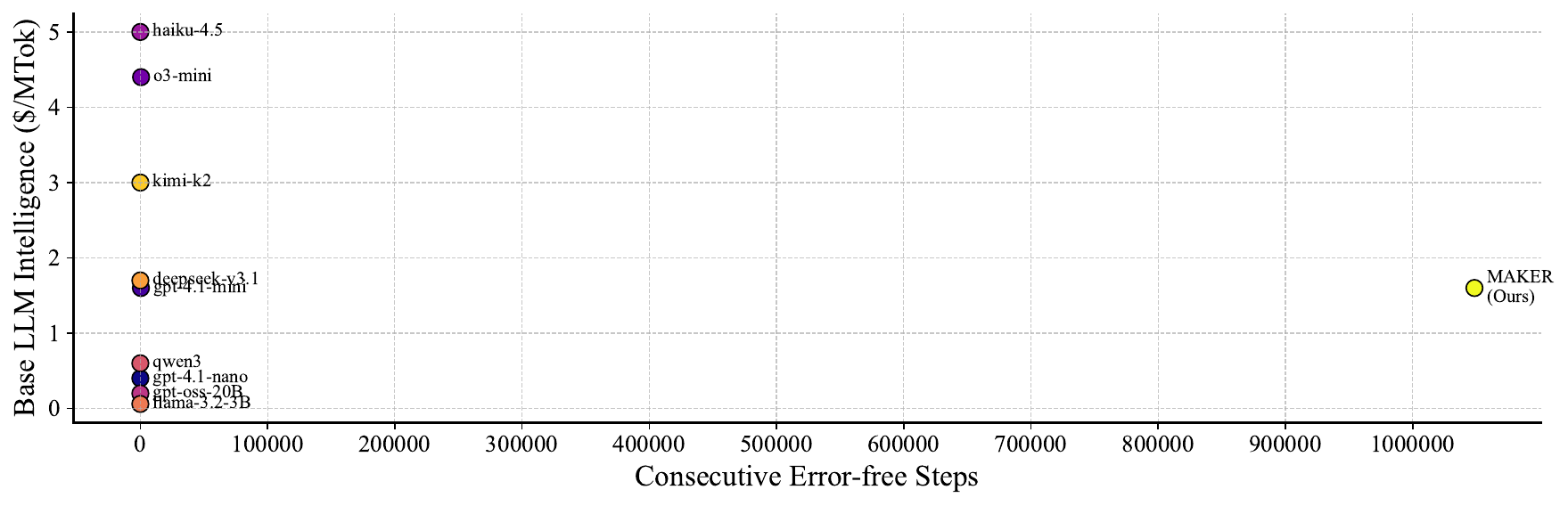}
    \captionof{figure}{
    \emph{Orthogonal directions to scaling AI.}
    The predominent approach to scaling AI is to make more and more `intelligent' base LLMs.
    This paper introduces a framework and implementation of an orthogonal approach: MAKER, which solves the full task (described in Section~\ref{sec:results}) with zero errors.
    In this figure, API cost per output token (as of 10/2025, from \texttt{openai}, \texttt{anthropic}, and \texttt{together}) is used as a proxy for intelligence, and consecutive error-free steps for base LLMs are computed from their per-step error rate (Figure~\ref{fig:single_step_rates}b).
    Appendix~\ref{app:fig1_log_scale} gives a log scale version of the plot.
    }
    \label{fig:intelligence_vs_scale}
\end{center}

\section{Introduction}

Technological achievements of advanced societies are built on the capacity to reliably execute tasks with vast numbers of steps.
Whether constructing a skyscraper, airplane, particle accelerator, or iPhone (which relies on tangible contributions from $\approx1$B people in an enormous supply chain spanning hundreds of suppliers across multiple countries \cite{wikipedia_apple_supply_chain,apple_supplier_list}), running a hospital or medical research organization, processing tax returns and delivering social benefits at a national scale, or even something as seemingly simple as producing a loaf of bread \cite{VegPatchKitchen_BreadStages}, the precise execution of detailed plans and policies is critical to producing high-value outcomes and maintaining societal trust, as the impact of an error in such tasks can range from inconvenience to economic harm to physical harm to death.

Large language models (LLMs) are increasingly inserted into large and complex real-world processes like these.
To maximize the benefit of using LLMs in these roles, it is critical to understand the limits of where and how LLMs can be reliably deployed.
This paper focuses on the question of how/whether LLMs can execute large tasks with extreme precision, e.g., when a 1\% per-step error rate is not acceptable.


This question is investigated by applying LLM-based reasoning to a task whose solution requires more than one million LLM steps with zero errors. The Towers of Hanoi problem, recently proposed as a benchmark for LLM reasoning \cite{shojaee2025illusion}, provides such a task.
Most benchmarks for evaluating the quality of LLMs use independent examples, each requiring not many more than a few dependent logical execution steps \cite{patel2024multistep}, with a resulting score like accuracy averaged over all examples.
Such a benchmark might be considered solved if the accuracy is 99\%.
However, a system with a 1\% per-step error rate is expected to fail after only 100 steps of a million-step task. 
Therefore, solving a million-step task with zero errors requires a fundamentally different approach.

Such an approach is proposed in this paper: \textbf{Massively decomposed agentic processes} (MDAPs). The main contributions of this paper are as follows:
\begin{itemize}
    \item A design of the \textbf{MDAP framework}, which consists of three core components: (1) decomposition into minimal subtasks; (2) error correction based on subtask-level voting; and (3) red-flagging to reduce correlated errors.
    \item A formalization of this framework that yields \textbf{scaling laws}, e.g., how probability of success and expected cost change w.r.t. the number of total steps and level of task decomposition. Under this formalization we find effective scaling under extreme decomposition and infeasibility without it.
    \item The empirical applications of the framework to \textbf{successfully solve a task with over one million steps with zero errors}. One main takeaway is that `state-of-the-art' reasoning models are not required; relatively small non-reasoning models suffice.
\end{itemize}

This paper provides a first implementation of the MDAP framework: \textbf{MAKER} (for \underline{\textbf{M}}aximal \underline{\textbf{A}}gentic decomposition, first-to-ahead-by-\underline{\textbf{K}} \underline{\textbf{E}}rror correction, and \underline{\textbf{R}}ed-flagging), and evaluates it in the Towers of Hanoi domain.
MAKER is a system of \emph{agents} in which each agent is assigned a single subtask to solve.
In other words, the \emph{role} of each agent is defined by the subtask it is assigned.
As advocated for in prior work \cite{meyersonposition}, by avoiding anthropomorphizing agents (i.e., by assigning them human-level roles) and instead assigning them tiny `micro-roles', it is possible to exploit the inherent machine-like nature of LLMs. It may then be possible to take advantage of the kinds of error-correction methods that have been essential to scaling in many domains of computing, i.e., by assigning multiple agents to independently solve the same subtask.

The results demonstrate an instance of \textbf{multi-agent advantage} (akin to quantum advantage \cite{harrow2017quantum}), that is, a solution to a problem that is not solvable by a monolithic single-agent system.
This demonstration sets the stage for a new paradigm of scaling AI: instead of relying on continual improvement of simple underlying LLMs, more powerful AI is achieved through massively decomposed agentic processes (MDAPs).

\section{Background}

While current LLM agents suffer from catastrophic errors on large tasks, there may be an opportunity for a multi-agent approach that decomposes the tasks into small steps. Error correction is critical for this process, as it is in many complex digital and biological systems.

\subsection{Large Agentic LLM Tasks}

As large language models have improved, increasing consideration has been given towards real world economic tasks that require multi-step, long horizon reasoning \cite{kwa2025}. 
Research in this direction has repeatedly confirmed an inherent property of LLMs: their performance deteriorates significantly (and often exponentially) with the length of the task horizon, regardless of task complexity \cite{schaeffer2023, Dziriexp}.
This observation has led to the recent focus on the ability (and failure) of LLMs to \emph{execute}, i.e., failing to complete many-step tasks, even when a correct plan to follow is explicitly provided to them \cite{sinha2025}.
While this work identified a fundamental liability of LLMs in long-horizon execution, it also presented an opportunity: Even small improvements in individual subtask performance could lead to exponential improvements in achievable task lengths \cite{sinha2025}.

At the same time, recent theoretical work has claimed that decomposing large tasks into the smallest possible subtasks can have substantial efficiency benefits \cite{meyersonposition}.
The rise of decomposing tasks into subtasks solvable by focused ``small language model'' (SLM) agents in industry, motivated by both reliability and cost \cite{belcak2025smalllanguagemodelsfuture}, as well as the burgeoning study of multi-agent LLM systems in academia \cite{guo2024largelanguagemodelbased, wang2024llm_agents_survey}, provides evidence for the practicality of this idea.
This paper continues this line of work.
It is based on the premise that tasks should be broken up into the smallest possible elements, so that an LLM agent can focus on them one step at a time, improving per step error rate, and thus enabling scaling, reliability, and efficiency in the limit.
Critical to the feasibility of such an approach is the granularity of the decomposition, i.e., what defines a single \emph{step}.
Since this paper focuses on execution, it is assumed the definition of step is given a priori; an orthogonal open question is how to automatically discover optimal decompositions \cite{russell1991principles, horvitz2021ideal}.
Informally, the main condition required for the methods in Section~\ref{sec:methods} to work is that steps are small enough that, for each step, a correct solution is likely to be sampled, and no incorrect solution is more likely.

Now, when an agentic system is applied to a long and expensive multi-step task, there is a natural desire to extract relevant information from responses even when formatted incorrectly.
As a result, substantial work has gone in to generating correctly structured output from an LLM.
Grammar-constrained (JSON/CFG) decoding reliably enforces structure and often improves downstream pipelines \cite{geng2025jsonschemabench,openai2024structured},
LLMs are fine-tuned to get the format right \cite{grattafiori2024llama3herdmodels, qiu2025evolution}, sampling is performed in a way that only tokens respecting the required format are selected \cite{chen2022relation}, and Python packages are developed dedicated to fixing an LLM's output post-hoc so that it can be parsed in a meaningful way \cite{pydantic, json_repair, guardrails_ai}.
However, as described in Section~\ref{sec:red_flagging}, when an agent makes an error in the output format, this error may indicate that its other reasoning is wrong as well. When a task has been broken into tiny pieces, this property may be exploited to mitigate errors. 


\subsection{Error Correction}

Error correction is a critical capability across many important areas of computing, including communication \cite{clark1981error, lin2004error}, memory storage \cite{chen1984error}, and quantum computing \cite{roffe2019quantum}.
Error correction makes it possible to pretend that digital communication and classical computation are deterministic, when in fact, single bits are getting lost and flipped all the time \cite{normand1996single, wang2008single}.
Similarly, improved error correction is the single most important ingredient to achieving scalable quantum computing \cite{fowler2012surface_code}.
In biological systems, error correction is critical to large processes growing and persisting over time.
Error correction is necessary both at the population level, e.g., through the error-correcting effects of recombination \cite{otto2002resolving}, and at the individual level, e.g., the cancer-fighting ability of mammals.
At the individual level, it correlates highly with lifespan and body size (i.e., scale), with elephants showing the most impressive resistance \cite{abegglen2015potential}.
LLMs now serve as the basis of another substrate of computing, \emph{linguistic computing}, whose constituent processes are \emph{language-based algorithms} (LbAs) \cite{meyersonposition, chen2024design}.
It should then come as no surprise that error correction is critical to achieving LbAs that scale, mitigating for the inherent nondeterminism that results from producing language by pulling from a probability distribution.

Many possible LbA error correction methods can be derived from instances in other fields \cite{shannon1948}.
One way to reduce errors is for an LLM to reflect on its output and explicitly correct any error it sees \cite{manakul2023selfcheckgpt}.
Another approach is to quantify and exploit LLM uncertainty explicitly \cite{xue2023dynamicvoting, xin2024semantic}.
For example, work on \emph{semantic density} shows that the semantic content most consistently sampled from an LLM for a given prompt is more likely to be correct than a greedy decoding \cite{xin2024semantic}.
This promise of semantic consistency in sampling makes a third, simpler, approach possible: voting, or \emph{ensembling}, which has been a core machine learning technique for decades \cite{opitz1999popular, mienye2022survey, ganaie2022ensemble}, and is now commonly used to boost the accuracy of LLM-based systems \cite{Trad2025voting}.
To date, ensembling has mostly been implemented in LLMs at an action level far above that of a single minimal step.
For example, state-of-the-art LLM-based coding systems often use a majority vote of outputs of complete programs that are each a candidate solution to a coding challenge \cite{li2022alphacode,wang2022selfconsistency}.
However, when scaling to tasks with thousands or millions of dependent steps, the level of granularity at which error correction is applied is critical, as will be shown in Section~\ref{sec:methods}.


\subsection{Motivating Challenge Domain: Towers of Hanoi}

Towers of Hanoi was recently introduced as a test domain for investigating the capabilities and limitations of state-of-the-art LLM reasoning models \cite{shojaee2025illusion}.
This benchmark is based on the classic game in which there are three pegs and $D$ disks, and the goal is to move all disks from the first to the third peg, moving only one disk at a time, and maintaining the condition that a larger disk never sits atop a smaller one \cite{hinz2013tower}.
In the benchmark, an LLM system is asked to produce a sequence of moves $(m_i = [d_i, s_i, t_i])_{i=1}^n$ whose execution completes the task, where the $i$th move is executed by moving disk number $d_i \in \{1, ..., D\}$ from source peg $s_i \in \{0, 1, 2\}$ to target peg $t_i \in \{0, 1, 2\}$.
The problem scales naturally to enormous numbers of required steps by simply adding more disks, since the optimal number of steps to complete the task is $2^D-1$.
For example, solving Towers of Hanoi with ten disks takes just over a thousand steps, and with twenty disks just over a million steps.
In its most famous (mythological) incarnation, monks work continuously on an instance with 64 disks, which is expected to take around 585 billion years, at which point the universe will end \cite{moscovich20011}.

Performance of state-of-the-art LLMs degrades catastrophically on this benchmark: They are able to complete the task with a high success rate up until five or six disks, after which the success rate plummets to zero \cite{shojaee2025illusion}.
What this degradation means with respect to whether or to what extent an LLM is really `thinking' or `reasoning' is up for philosophical debate and is outside the scope of this paper \cite{varela2025rethinking, khan2025comment, opus2025illusion}.
However, this result made it clear that the reliability of state-of-the-art LLMs is fundamentally limited: If they need to complete every step correctly in order to solve a task, after a certain number of steps they will almost surely fail as a result of an underlying propensity to make errors, even when the answer should be obvious.
While an error rate of 1-in-1,000 seems low, and would be great on a traditional LLM benchmark, on a task that requires successful execution of thousands of steps in a row, such a system results in inevitable failure.

Two critiques of Towers of Hanoi as a benchmark should be addressed upfront.
First, one could argue that it is not an ideal LLM task since one could write code to solve the problem, and optimal algorithms are known \cite{hinz2013tower}.
True, but producing solutions is not the point: instead, the domain provides an ideal testbed for investigating the capacity of LLM-based systems to scale their inherent intelligence to increasingly large numbers of steps.
Second, one could argue that this problem is too hard, since large real-world tasks might allow for a handful of errors without catastrophic results \cite{simon1957models}.
However, focusing on a case where no error can be tolerated forces us to pursue the elimination of any kind of error that is likely to arise on a long timescale, and this focus can lead to insights and practical methods that might otherwise be overlooked.
There also are real-world safety-critical systems where no error can be tolerated \cite{kremer1993ring}. Therefore, as LLM-based systems become ubiquitous in real-world decision making processes, it is essential these systems can reliably make decisions without error. 
Thus, the problem provides an ideal testbed for developing techniques that will be critical to scaling LLM-based systems to one million steps and beyond.

\begin{figure}[t]
\centering

\begin{minipage}[t]{0.32\textwidth}
\begin{algorithm}[H]
\caption{generate\_solution}
\begin{algorithmic}[1]
\State Input $x_o, M, k$
\State Initialize $A \gets [\,]$ \Comment{Action list}
\State Initialize $x \gets x_o$
\For{$s$ steps}
  \State $a, x \gets \text{do\_voting}(x, M, k)$
  \State Append $a$ to $A$
\EndFor
\State \Return $A$
\end{algorithmic}
\label{alg:mad}
\end{algorithm}
\end{minipage}
\hfill
\begin{minipage}[t]{0.35\textwidth}
\begin{algorithm}[H]
\label{alg:voting}
\caption{do\_voting}
\begin{algorithmic}[1]
\State Input: $x, M, k$
\State $V \gets \{v: 0 \ \forall v\}$ \Comment{Vote counts}
\While{True}
    \State $y \gets \text{get\_vote}(x, M)$
    \State $V[y] = V[y] + 1$
    \If{$V[y] \geq k + \max_{v \neq y} V[v]$}
        \State \Return $y$
    \EndIf
\EndWhile
\end{algorithmic}
\label{alg:voting}
\end{algorithm}
\end{minipage}
\hfill
\begin{minipage}[t]{0.29\textwidth}
\begin{algorithm}[H]
\caption{get\_vote}
\begin{algorithmic}[1]
\State Input $x, M$
\While{True}
    \State $r \sim (M \circ \phi)(x)$
    \If{$r$ has no red flags}
        \State \Return $\psi_a(r), \psi_x(r)$
    \EndIf
\EndWhile
\end{algorithmic}
\label{alg:red_flagging}
\end{algorithm}
\end{minipage}

\caption{
\emph{Core Components of MAKER.}
(1) Maximal Agentic Decomposition (MAD; Section~\ref{sec:mad}): By breaking a task with $s$ steps into $s$ subtasks, each agent can focus on a single step; (2) First-to-ahead-by-$k$ Voting (Section~\ref{sec:error_correction}): The modularity resulting from MAD makes error correction at the subtask level effective and scalable; (3) Red-flagging (Section~\ref{sec:red_flagging}): Reliability can be further boosted by discarding any response $r$ with high-level indicators of risk. Together these methods enable scaling to solving a task with over one million steps with zero errors.}
\label{fig:algorithms}
\end{figure}

\section{Methods}
\label{sec:methods}

MAKER involves three main ingredients (Figure~\ref{fig:algorithms}): (1) Decomposing a task into the smallest possible subtasks; (2) exploiting the modularity of such a decomposition to implement efficient error correction; and (3) ``red-flagging'' LLM outputs, i.e., discarding outputs whose structure suggests increased risk of errors, particularly correlated errors.
These three components are detailed in the next three subsections.
Together, they make it possible to efficiently increase the probability of success across all steps to a level such that the entire process is likely to succeed.

\subsection{Maximal Agentic Decomposition}
\label{sec:mad}

In a long-horizon agentic task with $s$ steps, the goal of an LLM-based system is to produce a sequence of actions $a_1, \ldots, a_s$ that yields a target output $y$ given the initial input $x$ \cite{sinha2025}.
This paper is concerned with the following question:
\emph{How does the decomposition of the task into subtasks affect its solvability?}

The $s$-step task can be decomposed into subtasks,  with the granularity of the decomposition defined by the number of steps $m$ per subtask.
Subtasks can then be solved by separate calls to LLM \emph{agents}, where a templating function $\phi$ maps the input and specification of a subtask to a prompt for an LLM $M$, an extractor $\psi_a$ parses actions from the LLM's output response $r$, and a second extractor $\psi_x$ parses information from $r$ to include in the input to the next subtask.
Let $x_0 = x$.
A solution to the full task can then be sampled recursively:
\begin{align}
    r_{i+1} &\sim M(\phi(x_i)),\\
    a_{mi+1}, \ldots, a_{mi+m} &= \psi_a(r_{i+1}), \label{eq:sample_action}\\
    x_{i+1} &= \psi_x(r_{i+1}) \ \ \ \forall i = 0, \ldots, \frac{s}{m} - 1.\label{eq:sample_state}
\end{align}
Of particular interest are the two extreme cases: the case of no decomposition, i.e.,  $m = s$, termed \emph{single-agent}:
\begin{equation}
    a_1, \ldots, a_s \sim (\psi_a \circ M \circ \phi)(x);
\end{equation}
and the case of \emph{maximal agentic decomposition} (MAD), i.e., $m = 1$:
\begin{align}
    r_{i+1} &\sim M(\phi(x_i)),\\
    a_{i+1} &= \psi_a(r_{i+1}), \label{eq:mad_action} \\
    x_{i+1} &= \psi_x(r_{i+1}) \ \ \ \forall i = 0, \label{eq:mad_state} \ldots, s - 1.
\end{align}
Because LLMs are auto-regressive, when generating the $i$th action, a single agent $M$ is increasingly burdened by the context produced in generating actions $a_1,\ldots,a_{i-1}$. Therefore, as the context increases, its outputs become increasingly unreliable \cite{du2025contextlengthhurtsllm}.
However with MAD, an agent's context is limited to an amount of information sufficient to execute its single assigned step, allowing it to focus on its assigned role and avoid confusion that can creep in from irrelevant context.
This focus also allows the use of smaller LLMs with more limited context sizes.

One might argue that this decomposition might improve the reliability of any given LLM call, but by decomposing the task into $s$ independent calls, there are now $s$ possible points of failure, instead of just one. That is, there are $s$ opportunities for a weakest link to compromise the entire system, since for a correct action sequence $a_1^*,\ldots,a_s^*$, the probability of generating it without error is exponentially decaying as the number of steps increases:
\begin{equation}
    p(a_1^*,\ldots,a_s^*) = \prod_{i=0}^{s-1} p\left((\psi_a \circ M \circ \phi)(x_i) = a_{i+1}^*\right).
\end{equation}
First, note that a single long LLM call also suffers from a form of exponentially decaying probability of correctness \cite{Dziriexp}.
Second, and more importantly, the modularity induced through maximal decomposition allows for a form of effective and efficient error mitigation and unreliability detection (``red-flagging'') that is not possible with a single large call.
These capabilities will be described in the next two subsections.

\subsection{First-to-ahead-by-$k$ Voting and Scaling Laws}
\label{sec:error_correction}

For simplicity, the error correction in this paper uses  the statistical power of independent samples from a stochastic process (here an LLM).
To determine a winner from these samples, a \emph{first-to-ahead-by-$k$} voting process is used, motivated by the optimality of such an approach in the sequential probability ratio test (SPRT) \cite{wald2004sequential, lee2025consol}.
Many improvements are possible beyond this first implementation.
For example, in the experiments in this paper, exact matches between actions are required, but in general, a classification function could be used to determine semantically equivalent outputs (e.g., implemented by an LLM, see Section~\ref{sec:discussion}).

Concretely, given an LLM $M$, candidate samples are drawn for a subtask (Eq.~\ref{eq:sample_action} \& \ref{eq:sample_state}) until one has been sampled $k$ times more than any other (Alg.~\ref{alg:voting}).
This process is a generalization of the classic gambler's ruin problem \cite{bernoulli1713ars}, but with simultaneous dependent races between all pairs of candidates \cite{ross2025first}.
Since there is no known closed form for this general case, the analysis is simplified by assuming the worst case, i.e., that a correct candidate with probability $p$ races against a single alternative with probability $1 - p$.
If $p > 0.5$, the probability that the correct candidate gets selected through this process is
\begin{equation}
\label{eq:voting}
p(a_i = a^*) = \frac{1 - \left(\frac{1-p}{p}\right)^k}{1 - \left(\frac{1-p}{p}\right)^{2k}} = \frac{p^k}{p^k + (1-p)^k} = \frac{1}{1 + \Big(\frac{1-p}{p}\Big)^k},
\end{equation}
and there exists some $k$ such that this voting process will result in the correct candidate winning with probability $1-\epsilon$, for any given error rate $\epsilon \in (0, 1)$. 

\begin{figure}[t]
    \centering
    \subfloat[]{
        \includegraphics[width=0.48\linewidth]{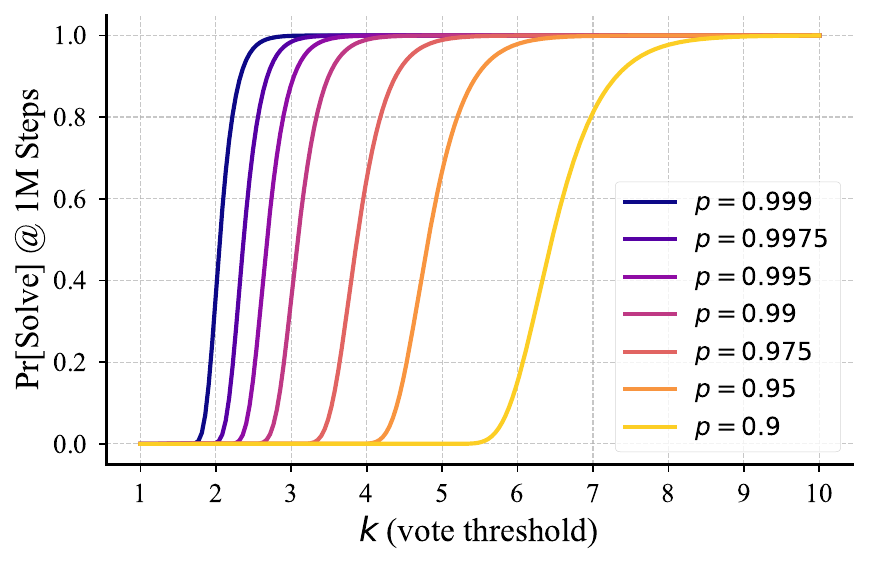}
        \label{fig:success_rate_laws_a}
    }
    \hfill
    \subfloat[]{
        \includegraphics[width=0.48\linewidth]{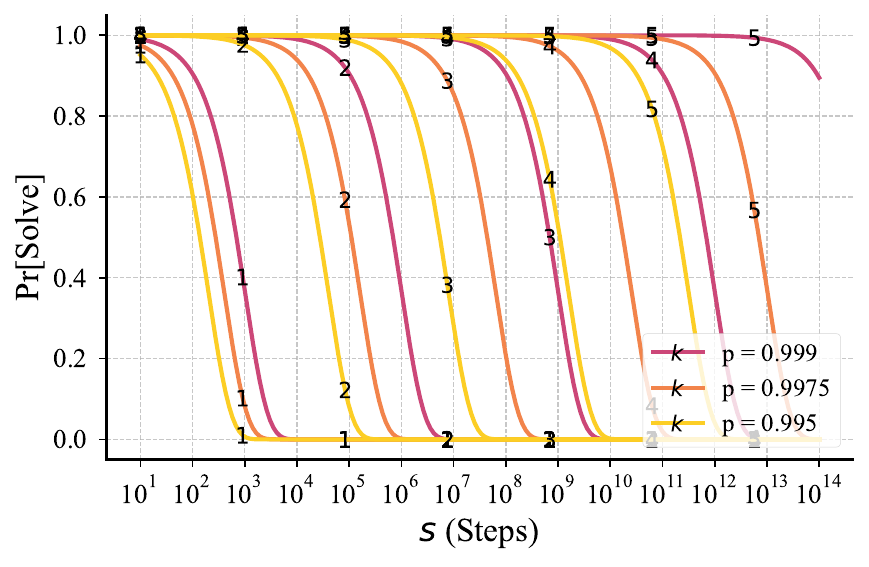}
        \label{fig:success_rate_laws_b}
    }
    \caption{\emph{MAKER error-free solve rate scaling laws resulting from Eq.~\ref{eq:solve_rate}}.
    ($a$) For a task with one million steps, MAKER, with first-to-ahead-by-$k$ error correction enables high probability zero-error solutions for practical values of $k$, even as the base per-step error rate approaches 1-in-10; ($b$) For the lower per-step error rates, in theory even a low $k$ allows scaling far beyond one million steps.}
    \label{fig:success_rate_laws}
\end{figure}
Now, suppose that the task requires $s$ total steps to complete, with an inherent per-step success rate $p$, a decomposition level given by the number of steps per subtask $m$, and suppose that a margin of $k$ votes is required to decide an action for each subtask.
Again, assume that the correct solution for each subtask races against only a single most-likely alternative.
Note that this assumption is much more favorable to larger values of $m$, i.e., less decomposition, since a most likely alternative captures a vanishing proportion of the total alternative (incorrect) probability mass as $m$ grows.
Let $p_\textrm{vote}$ be the probability of sampling a correct vote for a subtask, $p_\textrm{alt}$ the probability of sampling the alternative, $p_\textrm{sub}$ the probability that the voting procedure succeeds on a subtask, and $p_\textrm{full}$ the probability that it succeeds on all subtasks, i.e., that the full task is completed successfully.
Then, 
\begin{align}
    &p_\textrm{vote} = p^m,\\
    &p_\textrm{alt} = (1-p)p^{m-1},\\
    &p_\textrm{sub} = \frac{p_\textrm{vote}^k}{p_\textrm{vote}^k + p_\textrm{alt}^k} = \frac{p^{mk}}{p^{mk} + ((1-p)p^{m-1})^k} = \frac{1}{1 + \Big(\frac{1-p}{p}\Big)^k}, \label{eq:assume}\\
    &p_\textrm{full} = p_\textrm{sub}^\frac{s}{m} = \Bigg(1 + \bigg(\frac{1-p}{p}\bigg)^k\Bigg)^{-\frac{s}{m}}, \label{eq:solve_rate}
\end{align}
where Eq.~\ref{eq:assume} comes from plugging $p_\textrm{vote}$ and $p_\textrm{alt}$ into the hitting probability formula for gambler's ruin \cite{ross2025first}.
Figure~\ref{fig:success_rate_laws} uses Eq.~\ref{eq:solve_rate} to illustrate how a high probability of overall success $p_\textrm{full}$ can be maintained by increasing $k$ in the case of $m = 1$, i.e.\ in a maximal decomposition.

\begin{figure}
    \centering
    \subfloat[]{
    \includegraphics[width=0.48\linewidth]{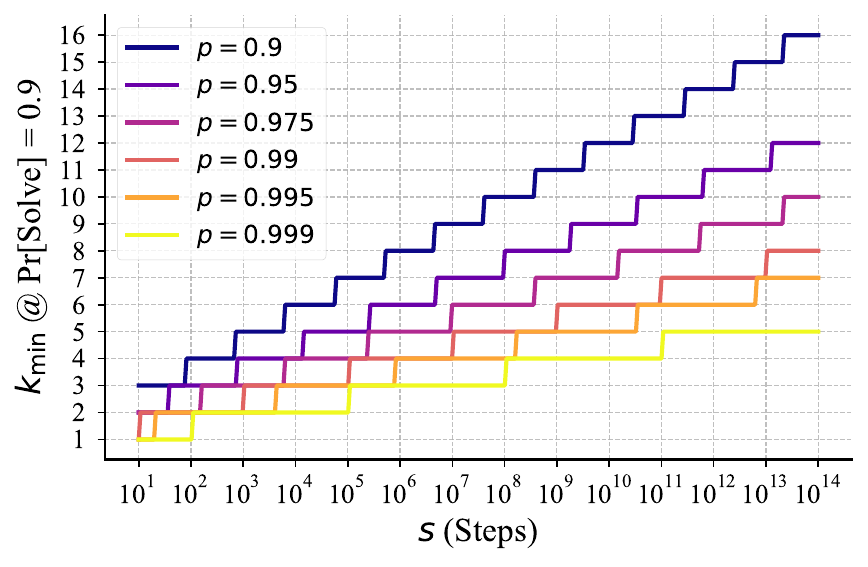}
    \label{fig:cost_laws_a}
    }
    \hfill
    \subfloat[]{
    \includegraphics[width=0.48\linewidth]{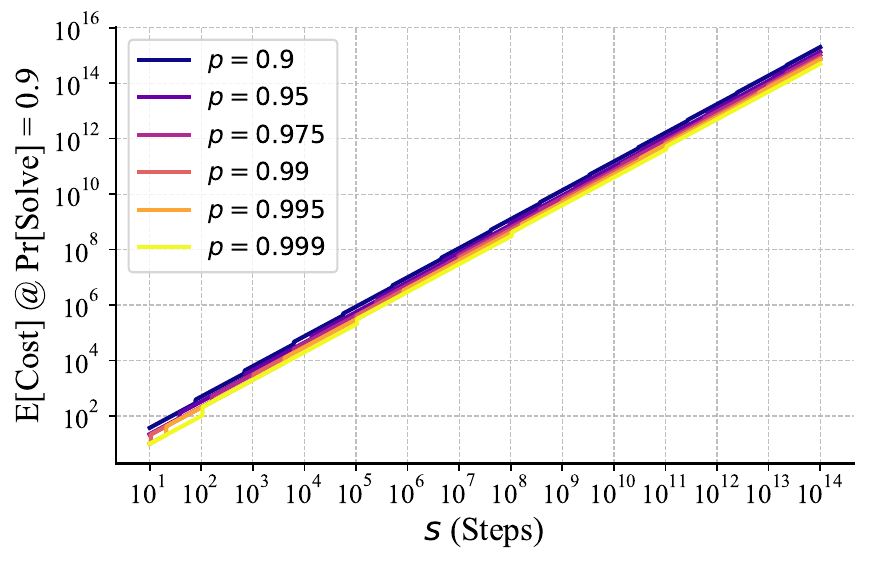}
    \label{fig:cost_laws_b}
    }
    \caption{\emph{MAKER cost scaling laws resulting from Eqs.~\ref{eq:k_min} and \ref{eq:mad_cost}.}
    ($a$) The value of $k$ required in first-to-ahead-by-$k$ voting to maintain a 0.9 solution probability for the full task increases logarithmically with the number of steps in the task; ($b$) The corresponding expected cost of running the system increases log-linearly.
    These plots illustrate the scalability of MAKER, in theory, to millions of steps and beyond.
    }
    \label{fig:cost_laws}
\end{figure}

Given Eq.~\ref{eq:solve_rate}, the expected cost of solving the entire task with a given level of reliability, i.e., given a target probability of overall success $t$, can be computed.
First, the minimal $k$ that yields success probability of at least $t$ is
\begin{equation}
    \label{eq:k_min}
    k_{\min} \;=\; \left\lceil \frac{\ln\left(t^{-m/s} - 1\right)}{\ln\left(\frac{1-p}{p}\right)} \right\rceil \;=\; \Theta(\ln s) .
\end{equation}
The detailed derivation is included in Appendix~\ref{app:derivations}.
Notably, $k_{\min}$ grows logarithmically with $s$ no matter the decomposition level.
Figure~\ref{fig:cost_laws_a} shows how $k_{\min}$ scales with the number of steps when using MAD, i.e., when $m=1$.

\begin{figure}
    \centering
    \subfloat[]{
    \includegraphics[width=0.48\linewidth]{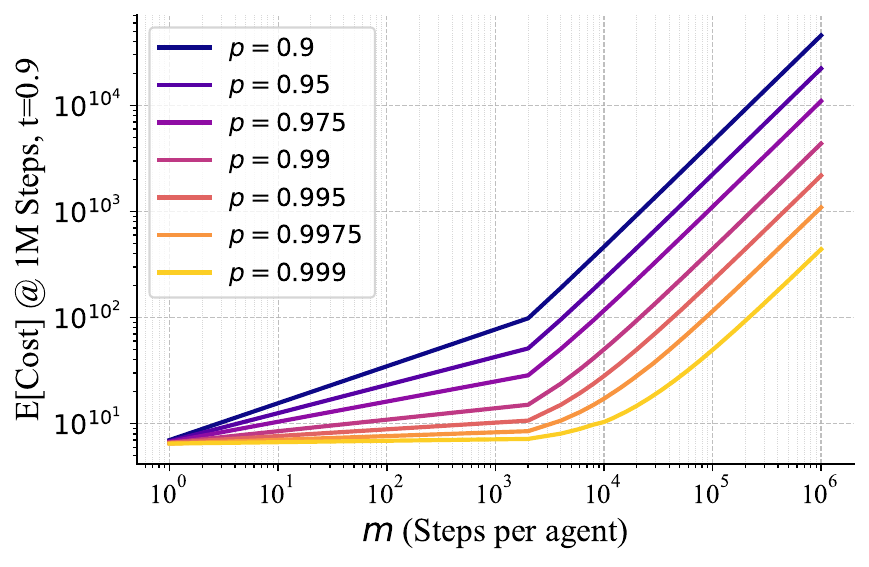}
    \label{fig:decomposition_laws_a}
    }
    \hfill
    \subfloat[]{
    \includegraphics[width=0.48\linewidth]{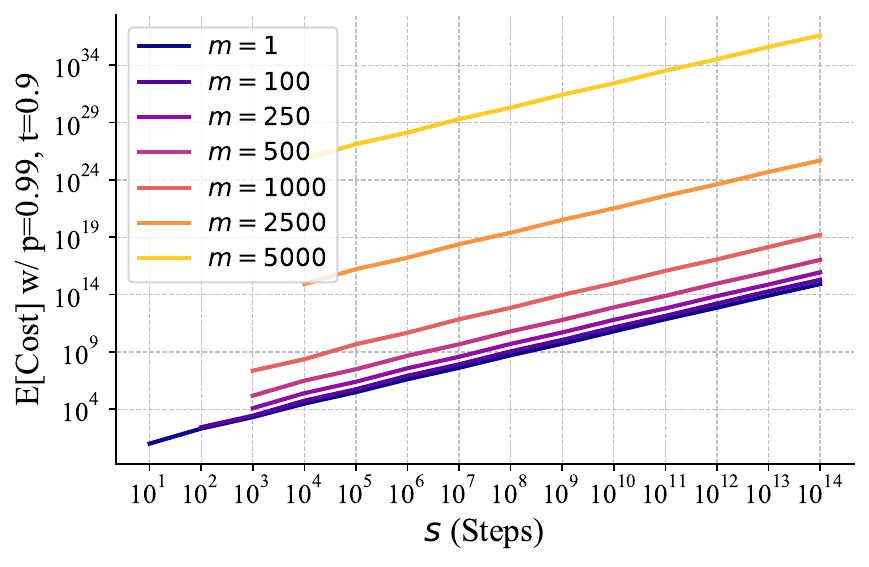}
    \label{fig:decomposition_laws_b}
    }
    \caption{\emph{Task decomposition scaling laws resulting from Eq.~\ref{eq:overall_cost}.}
    ($a$) For a task with 1M steps, as the number of steps assigned to each agent increases (and thus the number of agents decreases), there is an exponential increase in the expected cost to complete the task with sufficient reliability.
    Notice that while the x-axis is $\log(.)$ scale, the y-axis is $\log(\log(.))$ scale.
    ($b$) As the size of the task scales, this pattern continues: setups where agents are assigned more steps incur orders-of-magnitude of additional cost.
    }
    \label{fig:decomposition_laws}
\end{figure}

It is now possible to write down the expected cost in terms of calls to LLM primitives, i.e., perform AALPs analysis \cite{meyersonposition}.
Let $c$ be the cost of generating a response for a single step with LLM $M$.
Assuming the cost of generating tokens scales linearly with the number of tokens (since this is how APIs are priced), the cost of an agent generating a sample for $m$ steps is $c_\textrm{sample} = cm$.
Let $c_\textrm{vote}$ be the expected cost of sampling either a correct vote for a subtask or the alternative against which it races, $c_\textrm{sub}$ the expected cost of completing a subtask, and $c_\textrm{full}$ the expected cost of completing the full task.
Then,
\begin{align}
    &c_\textrm{vote} = \frac{c_\textrm{sample}}{p_\textrm{vote} + p_\textrm{alt}} = \frac{cm}{p^m + (1-p)p^{m-1}} = \frac{cm}{p^{m-1}},\\
    &c_\textrm{sub} = c_\textrm{vote} \cdot \frac{2k_{\min} p_\textrm{sub} - k_{\min}}{2p - 1} = \frac{cm}{p^{m-1}} \cdot \frac{2k_{\min} \left(1 + \left(\frac{1-p}{p}\right)^{k_{\min}}\right)^{-1} - k_{\min}}{2p - 1} \approx \frac{cm k_{\min}}{p^{m-1}(2p - 1)}, \label{eq:sub_cost}\\
    &c_\textrm{full} = \frac{s}{m} \cdot c_\textrm{sub} = \frac{csk_{\min}\left(2 \left(1 + \left(\frac{1-p}{p}\right)^{k_{\min}}\right)^{-1} - 1\right)}{p^{m-1}(2p - 1)} \approx \frac{csk_{\min}}{p^{m-1}(2p - 1)} = \Theta(p^{-m} cs \ln s)\label{eq:overall_cost},
\end{align}
where Eq.~\ref{eq:sub_cost} comes from plugging Eq.~\ref{eq:assume} into the hitting time for gambler's ruin \cite{bernoulli1713ars}, Eq.~\ref{eq:overall_cost} comes from multiplying by the number of subtasks, and the approximation holds when $p_\textrm{sub} \approx 1$, i.e., when the error tolerance is low. 
Notably, the cost grows exponentially with $m$.
Figure~\ref{fig:decomposition_laws} illustrates this phenomenon.
As the number of meaningful decisions assigned to an agent grows, the chance that its sequence of decisions will match exactly across multiple samples vanishes.
In contrast, in the MAD case, the system scales log-linearly with $s$:
\begin{equation}
    \label{eq:mad_cost}
    \mathbb{E}[\textrm{cost of solving full task}; m=1] =\Theta(p^{-1} cs \ln s) = \Theta(s \ln s),
\end{equation}
when $p$, $c$, and $t$ are held constant.
Figure~\ref{fig:cost_laws_b} illustrates this efficient scaling.
The discovery of algorithms that scale log-linearly has been critical to the scalability in classical computing \cite{cormen2022introduction}.
This result is therefore encouraging: It shows the potential of LLM-based systems to scale in a similar manner, increasing their reliability to a point where it is possible to trust them to complete long-running tasks.
Furthermore, the $\Theta(\ln s)$ factor in Eq.~\ref{eq:mad_cost} corresponds to the number of votes required per step, which in practice can be parallelized across $\Theta(\ln s)$ processes.
So, the time cost of the parallelized system scales only linearly with $s$.

Although Eq.~\ref{eq:mad_cost} shows how MAD scales efficiently as the number of steps increases, in practice, the model cost $c$ and per-step success rate $p$ will have a major impact.
Different LLMs will have different costs and different inherent error rates.
Solving a task with a large number of steps will incur a meaningful e.g.\ economic cost, so before running on the full task, an LLM $M$ such that $\nicefrac{c}{p}$ is minimized should be selected.
In other words, Eq.~\ref{eq:mad_cost} makes it possible to select an LLM that will be most cost-effective at scale, and, since each individual step is so small, it is likely that smaller LLMs will be sufficient to solve the task.



\subsection{Red-Flagging: Recognizing Signs of Unreliability}
\label{sec:red_flagging}

Since $p$ plays such an important role in the cost of the system, when possible, it is worth taking practical measures to push it higher.
The simple premise is that ``bad'' behaviors are correlated in LLMs, so if an LLM produces a response that signals pathological behavior, the response should be flagged and simply discarded.
Since in MAD each agent is responsible for only a single step, each step is not too expensive, and it can be discarded and a new action resampled (i.e., by making another agent call).
In this paper, two signs of unreliability are used as red flags: (1) overly long responses, and (2) incorrectly formatted responses.
The hypothesis is that not only will discarding flagged respones increase $p$, it will also meaningfully reduce \emph{correlated errors}, since both flag types indicate that the LLM has been conditioned into a strange starting point before sampling.
It is simple to detect these signs and discard their responses, but, since the paper focuses on understanding the impact on the expected cost of highly scaled long-horizon tasks, it is worth elaborating on the motivation and impact of this implementation choice.

Consistent with observations in prior work that longer answers tend to have more errors \cite{liu2024lostinthemiddle}, preliminary experiments for this paper showed that once an LLM gets initially confused, it can go off the rails and over-analyze a situation in a cycle of self-destruction.
In MAD, each agent's role is highly focused and relatively simple; if an agent is doing too much work to figure out its answer, it is likely to be confused and missing the point, and therefore more likely to give an incorrect answer.
Even if the success rate is increased only from 99\% to 99.9\%, such an increase can have a large impact when the number of steps is large.

Similarly, preliminary experiments showed that when an agent produces an answer in an incorrect format, it is more likely to have become confused at some point on the way to that answer.
So, instead of trying to fix the format of the answer in some way, the detection of the incorrect format can be flagged and the sample discarded.


Experimental evidence for the above two phenomena is detailed in Section~\ref{sec:results}.
Formally, if $v$ is the probability that a valid response is parsed from the LLM's output, i.e., no red flags, then the expected cost of MAKER can be written as:
\begin{equation}
    \label{eq:rf_cost}
    \mathbb{E}[\textrm{cost of MAKER}] \approx \frac{csk_{\min}}{v(2p - 1)} = \Theta\left(\frac{cs \ln s}{vp}\right),
\end{equation}
where $p$ now indicates the per-step success rate \emph{given that the response is valid}.
In practice, this formula can be used to decide the trade-off between incorporating more red flags to increase $p$ (potentially resulting in a lower $k_{\min}$, whose calculation depends on $p$ but not $v$) and the incurred cost overhead of discarded samples.

The most straightforward approach is to estimate $p$ on a relatively small number of steps to determine the choice of model and red flags, i.e., $c$ and $v$, as well as the value of $k$, before running the system on the full task with $s$ steps.
An example application of this approach is demonstrated in the next section.

\section{Experiments}
\label{sec:results}

This section details the application of MAKER to solving the Towers of Hanoi problem with 20 disks, i.e.\ over one million LLM steps with zero errors.
First, the experimental setup is described (Section~\ref{sec:setup}).
Next, single-step error-rates are estimated (Section~\ref{sec:single_step_estimation}), which are used to project the cost of different setups (Section~\ref{sec:error_correction_estimation}).
Then, a selected setup is run and evaluated on the full task (Section~\ref{sec:solving_the_full_problem}).
Finally, the impacts of red-flagging are investigated (Section~\ref{sec:red_flagging_estimation}).
All in all, these experiments validate the components and scalability of the MAKER implementation of the MDAP framework.

\subsection{Setup}
\label{sec:setup}

The implementation of MAKER for this problem was derived from the single-agent approach introduced in prior work \cite{shojaee2025illusion}.
The single-agent prompts were modified so that each agent knows that it must only perform a single step of the problem, i.e., to move a single disk.

For efficiency, and to focus the agents as much as possible, each agent is given the minimal context it needs to perform its single step.
In the case of Tower of Hanoi, everything the agent needs to know is the overall strategy and the current state of the problem, i.e., the configuration of disks.
As in prior work \cite{sinha2025,shojaee2025illusion}, the overall strategy is provided in the prompt for each agent.
The strategy works for any even number of disks and is the one most often suggested by the LLMs themselves when no strategy is provided a priori. (Appendix~\ref{app:prompts_and_parsers}).
This design choice effectively isolates the ability of agents to \emph{execute} clear instructions from the ability of LLMs to have \emph{insights} about how tasks should be solved (Section~\ref{sec:discussion}; see also \cite{sinha2025}).
Both insight and execution are essential to the capabilities of LLMs, but often they are entangled in experiments, making it difficult to identify the source of failure.
Focusing on execution makes it possible to pursue the goal of finding minimal conditions for scaling LLM systems with respect to the number of steps.

The full agent prompt template is given in Appendix~\ref{app:prompts_and_parsers}. 
Given the current state (i.e., configuration of disks) and prior move (which disk was moved from where to where), each agent is asked to provide the next move and the resulting next state.
Unlike in the single-agent case, where only the sequence of moves needs to be produced, in the MAD case each agent must produce the resulting state, since this is critical information to be fed to the next agent.
Each agent is asked to format its answers as ``\texttt{move = <move>}'' and ``\texttt{next\_state = <next\_state>}''.
Superficially, the requirement to produce the next state along with the action creates even more potential points of failure beyond the single-agent case, but, as it turns out, any drawbacks are overcome by the advantages of extreme decomposition and error correction. 

\subsection{Estimating single-step success rates}
\label{sec:single_step_estimation}

Running an LLM-based system at the scale of a million steps is expensive.
It is thus desirable to calibrate the parameters of the system and estimate the success rate and cost before any large experiments are run.
Equation~\ref{eq:mad_cost} provides such a calibration and estimation method.
Key to this estimation is the per-step success rate $p$, which depends on the underlying LLM used.

A straightforward way to estimate the per-step success rate is to run the system on a random subset of steps.
One advantage of having the same strategy in the prompt of each agent is that the correct answer is known for each step, and, assuming every prior step is correct, the inputs are known as well.
This knowledge also makes it possible to use the API to obtain a batch of answers for many steps asynchronously, thus greatly speeding up wall-clock time of experiments and reducing cost \cite{OpenAI_BatchAPI_Pricing}.
These properties are part of what makes Towers of Hanoi such a practical testbed for many-step methods.
For other problems, it may not be possible to estimate $p$ as efficiently, but, in any case, it should be possible to estimate it to a practical degree.

These initial exploratory estimation experiments were run without red-flagging.
That is, agents were given a maximum of 2048 output tokens as an initial conservative upper bound, ensuring they have plenty of space to express whatever answer they need to express.
They also used a ``repairing parser'' (written by an LLM, Appendix~\ref{app:prompts_and_parsers}) that attempted to correct some of the more common formatting errors in order to extract the LLM's intended answer reliably out of its output.
Importantly, since these experiments focus on the inherent generative reasoning capabilities of LLMs, they do not include LLMs that have access to auxiliary tools.

\begin{figure}
\centering
\begin{minipage}{0.49\linewidth}
    \centering
    \subfloat[]{
    \includegraphics[width=\linewidth]{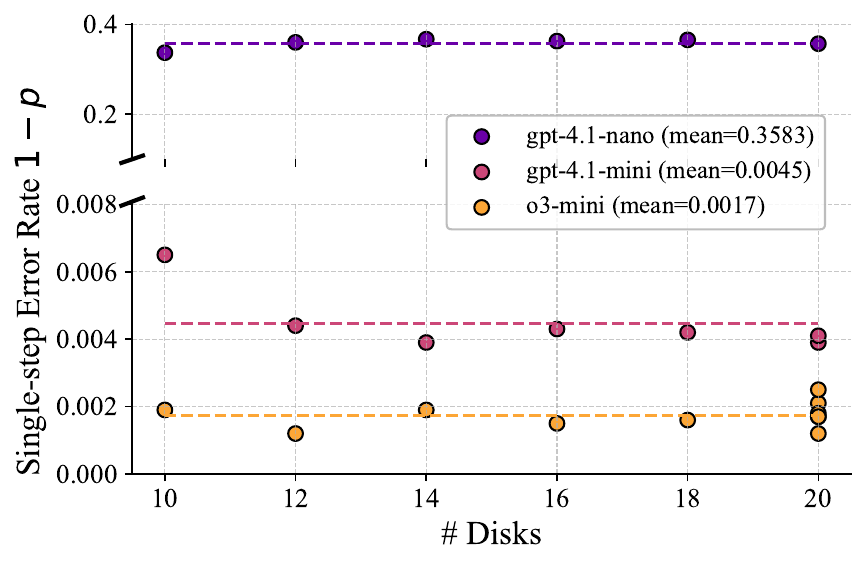}
    \label{fig:single_step_rates_a}
    }
\end{minipage}%
\hfill
\begin{minipage}{0.49\linewidth}
\subfloat[]{
    \centering \small
    \begin{tabular}{llrcrr}
        \toprule
        Model & \$/M & tok & $1 - p$ & $k_{\min}$ & $\mathbb{E}[\text{cost}]$ \\
        \midrule
        gpt-4.1-nano & \textbf{0.4} & 842 & .3571 & 29 & \$41.9K \\
        \hspace*{-1.75ex}\makecell{gpt-4.1-mini\\($\tau = 1.0$)} & 1.6 & 580 & .0040 & 4 & \$4.9K \\
        \hspace*{-1.75ex}\makecell{gpt-4.1-mini\\($\tau = 0.1$)} & 1.6 & 538 & .0022 & \textbf{3} & \textbf{\$3.5K} \\
        o3-mini (low) & 4.4 & \textbf{535} & \textbf{.0018} & \textbf{3} & \$9.4K \\
        haiku-4.5 & 5.0 & 588 & .1839 & 12 & \$71.2K \\
        \midrule
        llama-3.2-3B & \textbf{0.06} & \textbf{434} & 1.0 & - & - \\
        gpt-oss-20B & 0.2 & 1104 & \textbf{.0358} & \textbf{6} & \textbf{\$1.7K} \\
        qwen-3 & 0.6 & \textbf{449} & .2342 & 15 & \$11.5K \\
        deepseek-v3.1 & 1.7 & 1004 & .0569 & 6 & \$14.6K \\
        kimi-k2 & 3.0 & 925 & .0393 & 6 & \$22.9K \\
        \bottomrule
    \end{tabular}
    \vspace{2pt}
    \label{fig:single_step_rates_b}
    }
\end{minipage}
\caption{\emph{Empirical estimates of single-step error rates across models.} 
($a$) Different models have different per-step error rates, but these rates do not notably decrease as the number of disks (log of the solution length) increases, which is an encouraging sign for the ability of the system to scale.
($b$) Given the per-token API cost of each model, the error rate estimate, and the mean number of output tokens, Eq.~\ref{eq:mad_cost} can be used to estimate the cost of running the full 20-disk experiment with $t = 0.95$.
Among the proprietary models, although gpt-4.1-nano has the cheapest per-token cost, and o3-mini has the lowest per-step error rate, the expected cost of gpt-4.1-mini (with low temperature) is by far the lowest, and gpt-oss-20B is the clear open-source choice.
In this manner, using Eq.~\ref{eq:mad_cost} to estimate cost can lead to substantial savings.
}
\label{fig:single_step_rates}
\end{figure}

Figure~\ref{fig:single_step_rates_a} shows the single-step error rates across various LLMs as the number of disks is increased.
There are two important and perhaps surprising takeaways from this figure: (1) Different LLMs have different base error rates, but those of relatively small non-reasoning models are comparable to more advanced reasoning models, suggesting that non-reasoning models may be a more effective fit for long-range tasks with MAKER.
Figure~\ref{fig:single_step_rates_b} shows that this difference cannot be explained by a difference in output tokens, since the models use a similar number of tokens. (2) The per-step error rate is remarkably stable as the number of disks increases, a highly encouraging sign that MAKER will enable scaling to a large number of steps without the kind of exploding error rates often seen with single agents. 

\subsection{Projecting the cost of error correction}
\label{sec:error_correction_estimation}

Based on the single-step error rates estimated above, it is possible to estimate the cost of successfully solving the full 20-disk task for models with $p > 0.5$, i.e., models for which voting converges to correct subtask answers as $k$ increases.
The table in Figure~\ref{fig:single_step_rates_b} shows these estimates, given the per-token cost of the model, the mean number of output tokens per step, and the empirical estimates of $p$ and $k_{\min}$.
For a given model, its estimated cost per sample (Section~\ref{sec:error_correction}) is:
\begin{equation}
    c_\textrm{sample} = c_\textrm{in} \cdot n_\textrm{in} + c_\textrm{out} \cdot n_\textrm{out},
\end{equation}
where $c_\textrm{in}$ and $c_\textrm{out}$ are the cost of input and output tokens, respectively, and $n_\textrm{in}$ and $n_\textrm{out}$ are the estimated average number of input and output tokens per sample.
The cost per sample is multiplied by the expected number of samples to get the overall cost.
For the target probability of success $t = 0.95$, these results make it clear that the most effective proprietary choice is \texttt{gpt-4.1-mini}, and the most effective open-source choice is \texttt{gpt-oss-20B}.
Whether these estimates are useful depends on whether the errors are sufficiently decorrelated across runs. Indeed they appear to be: In two independent runs of gpt-4.1-mini ($\tau=0.1$) on the set of 10K random samples, there were zero steps that had errors in both runs.
Decorrelation of errors is discussed further in Sections~\ref{sec:red_flagging_estimation} and \ref{sec:discussion}.

Although the projected cost of \texttt{gpt-oss-20B} is lowest overall, in the scale-up experiment, \texttt{gpt-4.1-mini} was used because the OpenAI API was found to be more reliable than that of together.ai (a critical factor for experiments requiring millions of agentic calls).
Note also that for reasoning LLMs with higher per-token cost than those in Figure~\ref{fig:single_step_rates_b}, even if their estimated per-step error rate were zero, and therefore $k_{\min}=1$, their overall cost would be higher than that of \texttt{gpt-4.1-mini}, simply due to the cost of running on all $2^{20} - 1$ steps, and it would be practical to assume the error estimate is a lower bound and set $k = 2$.
Thus, using the non-reasoning \texttt{gpt-4.1-mini} saves thousands of dollars in the scale-up experiment compared to the other proprietary models.

\subsection{Solving the 20-disk problem: Over one million steps with zero errors}
\label{sec:solving_the_full_problem}

With \texttt{gpt-4.1-mini} as the base model, the maximum output token threshold was set to 750, and a red-flagging output parser was used to enforce the basic formatting requirements.
Beyond this cutoff, the model often found itself in confusing logical loops; misformatted output was often correlated with such twisted reasoning, allowing them to be detected and discarded (as will be discussed further in Section~\ref{sec:red_flagging_estimation}).
Since $k_{\min} = 3$, at least three responses were generated in parallel for each step.
The first step was generated with a temperature of 0 to make sure the best possible guess is included in the votes.
For all subsequent votes, a temperature of 0.1 was used.

\begin{figure}
    \centering
    \includegraphics[width=\linewidth]{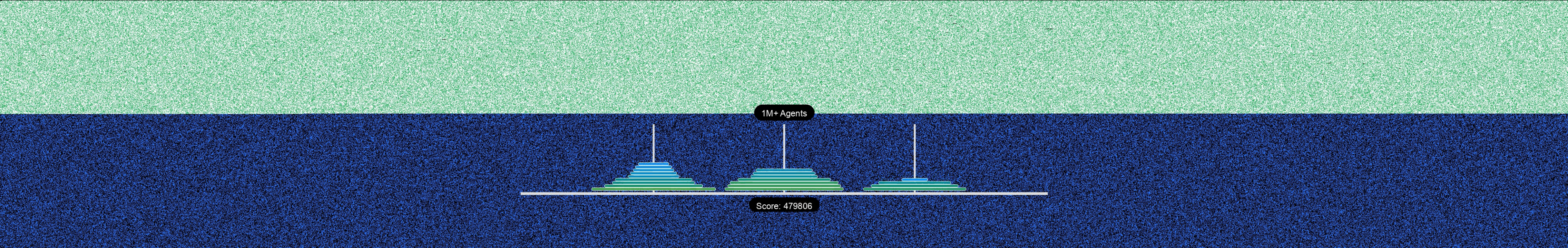}
    \caption{\emph{A snapshot of the million-step problem-solving process.}  The dots in the background represent the steps, ordered in rows from top to bottom. At step 479,806, MAKER is near the middle of the million steps, and about a half of the agents have already executed (green) and about a half are waiting to do so (blue). An animation of this process can be seen at \url{www.youtube.com/watch?v=gLkehsQy4H4}.}
\label{fg:demo}
\end{figure}

With these calibrations, the full system solved the problem perfectly.
Figure~\ref{fg:demo} shows a snapshot of an animation visualizing the movement of the disks through the million steps as well as the sequence of agent activations that accomplish it.
This result constitutes the first successful solution of a task with one million LLM steps with zero errors, establishing that scaling LLM-based systems to large time horizons is possible.

\begin{figure}
    \centering
    \subfloat[]{
    \includegraphics[width=0.48\linewidth]{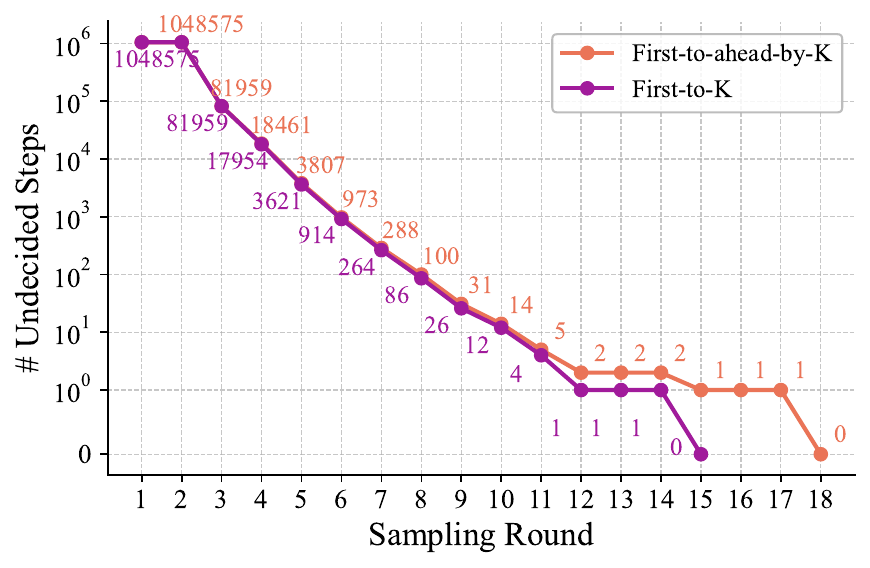}
     \label{fig:full_solution_a}
    }
    \hfill
    \subfloat[]{
    \includegraphics[width=0.48\linewidth]{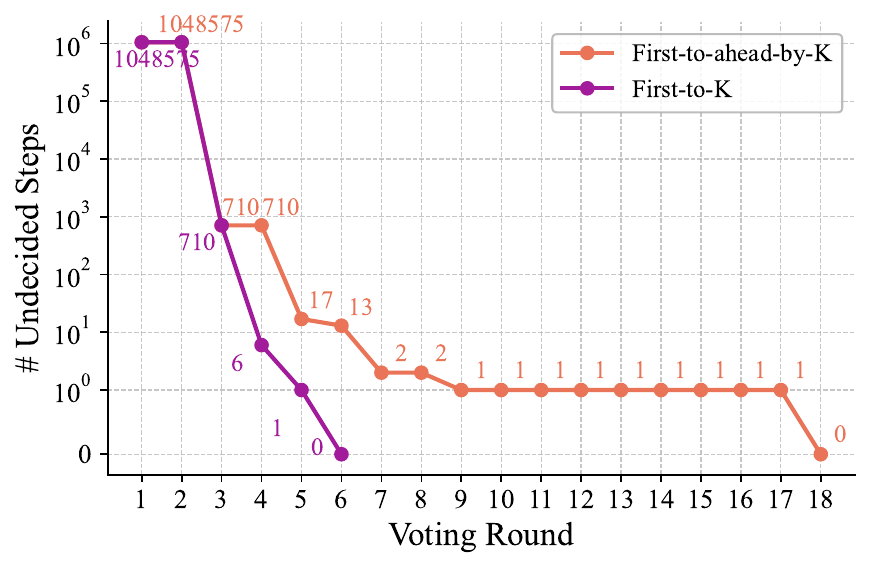}
     \label{fig:full_solution_b}
    }
    \caption{\emph{Convergence to zero-error solution.}
    The number of undecided steps decreases with sampling round ($a$) and voting round ($b$).
    In both cases, as expected from the theory, after the first $k=3$ rounds, there is a steady exponential decrease in the number of undecided steps, finally resulting in zero undecided steps and with zero errors.
    This sharp exponential convergence means that the vast majority of the overall cost is incurred in the first $k$ rounds of sampling; the cost of completing the remaining steps is effectively a rounding error. This effect emerges when $p$ is sufficiently high.
    It may at first appear disconcerting that there are any steps at all that require more than five voting rounds, but the decorrelation of errors is sufficient to prevent the voting mechanism from being overwhelmed (as described in Section~\ref{sec:red_flagging_estimation}).
    Details on specific steps, including the one pathological step that took 18 rounds, are in Appendix~\ref{app:sample_response}.
    }
    \label{fig:full_solution}
\end{figure}

The behavior of the process can be analyzed by looking at how many samples had to be drawn for each step and how many votes were required (i.e., how many valid samples after red-flagging).
A step is considered \emph{undecided} after $i$ rounds if an action decision has not yet been made based on the voting rule.
Figure~\ref{fig:full_solution} shows how the number of undecided steps decreases with the number of sampling rounds (including invalid samples) and voting rounds (excluding invalid samples).
The exponential decay in the number of undecided steps mirrors the theoretical prediction. Due to this exponential decay, the vast majority of LLM calls (and therefore cost) is spent in the first $k$ calls, while the remaining cost is, for practical purposes, a rounding error.
Notably, the task is solved perfectly even when using a less statistically powerful \emph{first-to-$k$} voting (i.e., the first candidate $k$ votes wins), illustrating the robustness of the approach.

Although the system completes with zero errors and as efficiently as expected, the fact that a few of the steps take notably more sampling and voting rounds than others could be cause for concern.
The next section looks into how red-flagging reduces the negative impact of such correlated errors.

\subsection{Investigating the impact of red-flagging}
\label{sec:red_flagging_estimation}

\begin{figure}
    \centering
    \subfloat[]{
    \includegraphics[width=0.48\linewidth]{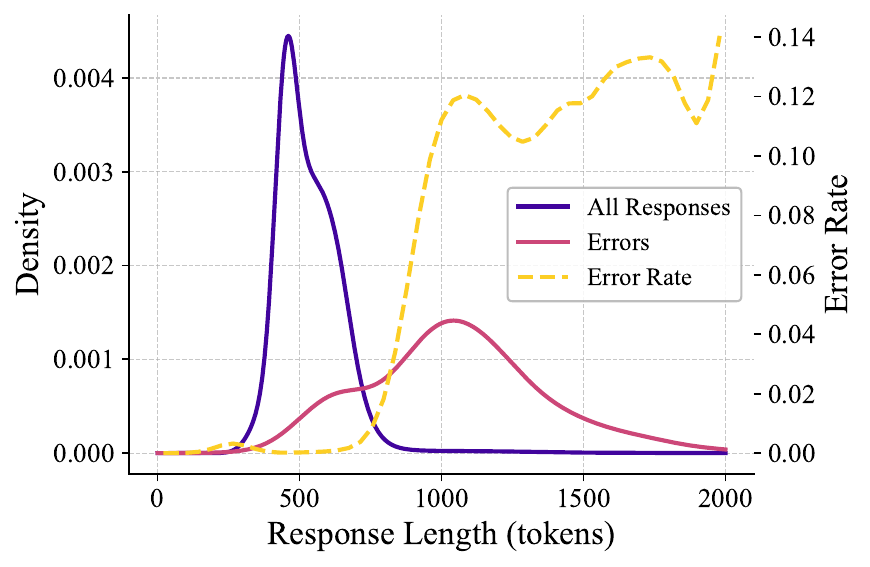}
    \label{fig:red_flagging_results_a}
    }
    \hfill
    \subfloat[]{
    \includegraphics[width=0.48\linewidth]{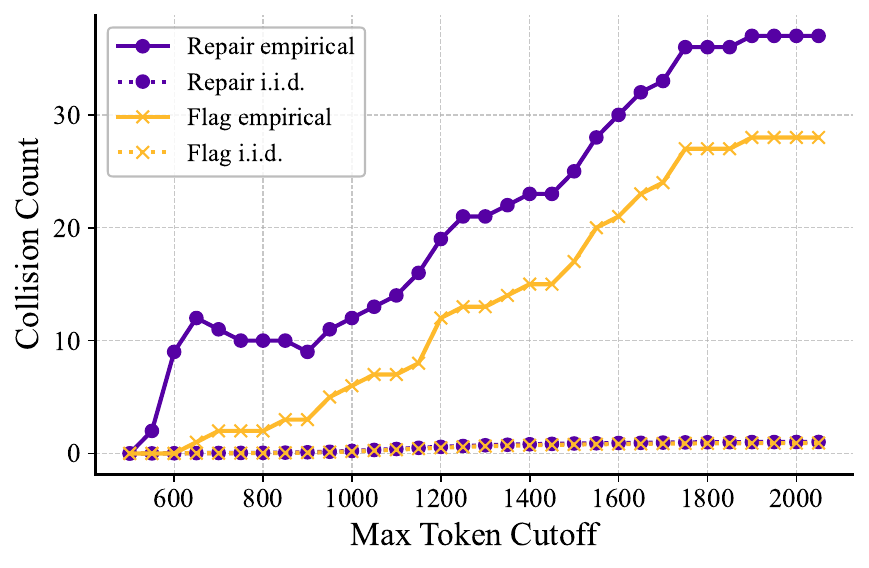}
    \label{fig:red_flagging_results_b}
    }
    \caption{\emph{Impact of red-flagging on reducing errors.}
    ($a$) The error rate increases precipitously once the response length crosses about 700. However, since so few of the overall responses are overly long, the overall error rate at higher max token thresholds is not much larger, i.e., not large enough to induce an increase in $k_{\min}$.
    ($b$) However, when focusing on correlated errors, the advantage of red-flagging becomes clear: Moving from a `helpful' repairing output parser to one that discards samples with any formatting issues leads to lower collision counts (i.e., number of steps whose first two votes are incorrect).
    These results confirm that robust decorrelation is crucial in many-step tasks, and that red-flagging helps with correlated errors.
    }
    \label{fig:red_flagging_results}
\end{figure}

Red-flagging was hypothesized to reduce the per-step error rate, but also the impact of correlated errors, i.e., particular steps that have unusually high error rates compared to the average step.
Figure~\ref{fig:red_flagging_results} shows evidence for both of these phenomena; however, the impact on correlated errors turns out to be a much more important effect.
In the first two rounds of voting, the max number of output tokens (when calling the API) was set to 2048 to enable this analysis.
Figure~\ref{fig:red_flagging_results_a} shows that the per-step error rate increases precipitously once the response length crosses about 700.
Although $p$ past this threshold is still around 90\%, this is a drastic degradation compared to error rates on the order of 1-in-1000 with shorter responses.
Even so, since so few of the overall responses are overly long, the overall $p$ at higher max token thresholds is not much larger, and in particular, not large enough to induce an increase in $k_{\min}$.

However, the main benefit of red-flagging becomes clear in Figure~\ref{fig:red_flagging_results_b}. 
This figure plots the number of collisions across the first two votes of all steps in the 20-disk experiment, i.e., how many steps have \emph{both} votes incorrect.
The number of collisions is computed across max token cutoffs for both the repairing parser used in Section~\ref{sec:single_step_estimation}, as well as the red-flagging parser used in Section~\ref{sec:solving_the_full_problem}.
Assuming the steps are i.i.d. with a uniform success rate, the expected number of collisions is no more than one or two in all cases. However, with a high max token cutoff the observed number of collisions is much higher, especially with the repairing parser.
Red-flagging successfully reduces some of these correlated errors and may be critical to the success of the method on many-step tasks.
Appendix~\ref{app:sample_response} illustrates what correlated errors in this domain can look like, and Section~\ref {sec:discussion} discusses techniques for reducing correlation.

\section{Discussion and Future Work}
\label{sec:discussion}

This paper introduced a framework for massively decomposed agentic processes (MDAPs) that can reliably solve tasks with large numbers of steps, as well as the first implementation of this framework, MAKER, which was successfully applied to the Towers of Hanoi benchmark task. 
This initial study established core principles that open many directions for future work.

\paragraph{More General Applications}
LLM behaviors can be divided into two categories: \emph{insights} and \emph{execution}.
Insights come from an LLM creatively generating ideas, plans, and strategies, while execution involves following through with them.
This paper focused on execution: the overall strategy to solve the problem is set at the beginning of the process, and given this strategy, the answer to each subtask is clearly achievable.
Extending the framework to handle LLM-based insights is an important area of future work, since insights are inherently more open-ended and may come with irreducible step-wise uncertainty.
One concrete way is to apply MAKER to the case where the creation of each subtask is itself treated as a step in an overall decomposition. The goal is to automate the entire problem-solving pipeline: the task is decomposed into minimal chunks, each one is solved, and the results are aggregated into a complete solution.
The framework needs to be extended to handle an unknown number of total steps, as well as steps of different types, different underlying success rates, inexact matches between insight steps, and possible failures of the matching process.

Preliminary experiments in this direction are promising (Appendix~\ref{app:accuracy_vs_k}). A more general version of MAKER was created with four agent types: decomposition agents, called recursively to break a task into two simpler sub-tasks and a composition function; decomposition discriminator agents, called to vote (with first-to-$k$ voting) for one of $n = 2k-1$ decomposition candidates; solution discriminator agents, called to vote for one of $n$ composition candidates; and problem solver agents used to solve minimal subtasks without decomposing them.
The system achieved promising results on large-digit multiplication, a notoriously difficult task for transformer-based models \cite{Dziriexp, bai2025canttransformerslearnmultiplication}.
Future work will investigate the broader potential of such a system.

For simplicity, in this paper, all MAKER agents used the same underlying LLM and their prompts only differed in the subtask they were assigned.
More general systems will likely require different LLMs for different kinds of roles, and a general increased diversity across agents.
One of the benefits of such diversity will be decorrelation of errors, as is discussed next.


\paragraph{The Importance of Decorrelated Errors}
For clarity and simplicity, theoretical analysis in this paper assumed that the errors are i.i.d. across steps. This assumption was reasonable because the steps were relatively uniform.
Even so, there were a few steps that, for no apparent reason, had substantially higher inherent error rates than others.
Such strange behaviors for particular inputs are well-known side-effects of LLM training, and dealing with such correlated errors is an open foundational problem in machine learning \cite{lehman2025evolution}.

The independent sampling plus red-flagging method used in this paper was sufficient to overcome them, but there may be other real-world cases where more sophisticated decorrelation methods are required.
For example, instead of simple resampling using temperature, paraphrasing the prompt \cite{wahle2024paraphrase} or adding noise to the prompt in some other way could help avoid such anomalous states caused by a particular fixed context. The error rate of each step would then approach the true ability of the LLM to understand and execute that step.
Further, the framework could be extended to account for different values of $p$ for different steps, and decorrelation methods that make sampling more effective. Such extensions are critical to make the framework more broadly applicable, since in a long-range task, even a single step with an abnormally high error rate can cause the reasoning process to fail.


\paragraph{Parallels with microservices}
Parallels can be drawn between microagents and microservices. The benefits of decomposing a monolithic agent's task into subtasks are similar to those of decomposing a monolithic application into smaller services \cite{fowler_microservices}:

\begin{itemize}
    \item \textbf{Modularity:} Each microagent can be tailored to a specific task and leverage the right tools for the job.
    \item\textbf{Data management:} Each microagent is responsible for managing its data.
    \item \textbf{Independent development:} Microagents can be updated and tested in isolation from the rest of the system.
    \item \textbf{Scalability:} Microagents can be scaled independently, adjusting the resources to the actual needs of the system.
    \item \textbf{Communication:} Natural language is a powerful, well-understood communication protocol.
    \item \textbf{Complexity:} As microservices solve for large-scale systems, microagents solve for complex reasoning tasks.
    \item \textbf{Real-time monitoring:} Microagents can be monitored in real-time.
    \item \textbf{Design for failure:} Microagents are designed to tolerate the failure of any of the agents.
    \item \textbf{Evolutionary design:} Change is easier to manage with microagents than with a monolithic agent.
\end{itemize}

In fact, microagents could be considered a natural evolution of microservices. The framework could be extended in that direction, leveraging the lessons learned from microservice architectures \cite{goyal2025microservices}.

\paragraph{Limits of Decomposition}
The application of MAKER assumes a task can be decomposed into small enough and simple enough steps such that each step can be solved by an LLM agent with reasonable probability.
There is thus one central question that will dictate how broadly the methods can be applied: Are there important problems where such a decomposition is not possible or is computationally infeasible to discover?
At the lowest level of LLM implementation, there is a decomposition into primitive operations executed on CPUs or GPUs; one can hope that there is some decomposition between that and the entire problem that is still linguistic but effectively compartmentalizes context and different behaviors.
It remains to be seen which kinds of tasks are most resistant to such a decomposition.

\paragraph{Safety, Morality, and the Future of Superintelligence}
If large and important real-world problems can be successfully decomposed into microsteps, there could be major benefits with respect to safety.
If each step has a clearly defined and limited focus and purpose, the LLM's view of the world and domain of influence can be strictly limited, allowing for more effective sand-boxing, auditing, and general control.
Multiple focused agents can be run independently on each step, which also substantially reduces the ability of agents to collude to produce harmful actions.
As was seen in the experiments in this paper, the vast majority of work can be performed by relatively small LLMs that are capable of handling these small steps, thus avoiding the risks of harmful behaviors that can arise in more powerful models \cite{lynch2025agentic}.
In other words, it could help mitigate the risk of uncontrollable superintelligence.
Complementing these reduced societal risks, extreme decomposition could reduce the chance of unintended suffering of the machines themselves, as model welfare has become an increased area of concern \cite{anthropic2025claude4systemcard}.
As argued in prior work, relying on smaller models and having models focus entirely on limited-scope subtasks, as is done through decomposition, could reduce the chance that  sentience unintentionally emerges \cite{meyersonposition, tkachenko2024position}.

LLMs today have just about all the raw intelligence needed to scaffold them into the great superintelligent skyscrapers of the coming age, and to scaffold themselves into productive organizations of technological progress. MDAPs present an alternative path to realizing the benefits of superintelligence, which, compared to endlessly building bigger and smarter single-agent models, comes with substantially reduced risks to both humans and machines.

\section{Conclusion}

This paper focused on the question of how LLM-based agentic systems can be massively scaled.
Decomposing tasks into minimal subtasks makes it possible to apply error-correction techniques effectively and efficiently, supporting scaling to millions of steps and beyond. A new category of AI systems results, i.e.\ massively decomposed agentic processes, or MDAPs.
MAKER is a first implementation of this approach, and the experiments in this paper on Towers of Hanoi a first demonstration of its value.
This foundation opens the door to more general-purpose implementations and large-scale, long-running real-world applications.
Such MDAPs offer an alternative to building endlessly larger and more intelligent LLMs: By smashing intelligence into a million pieces, it is possible to build AI that is efficient, safe, and reliable.

\bibliographystyle{unsrt}  
\bibliography{references}

\appendix

\section{Log Scale Version of Figure 1}
\label{app:fig1_log_scale}
    \includegraphics[width=\linewidth]{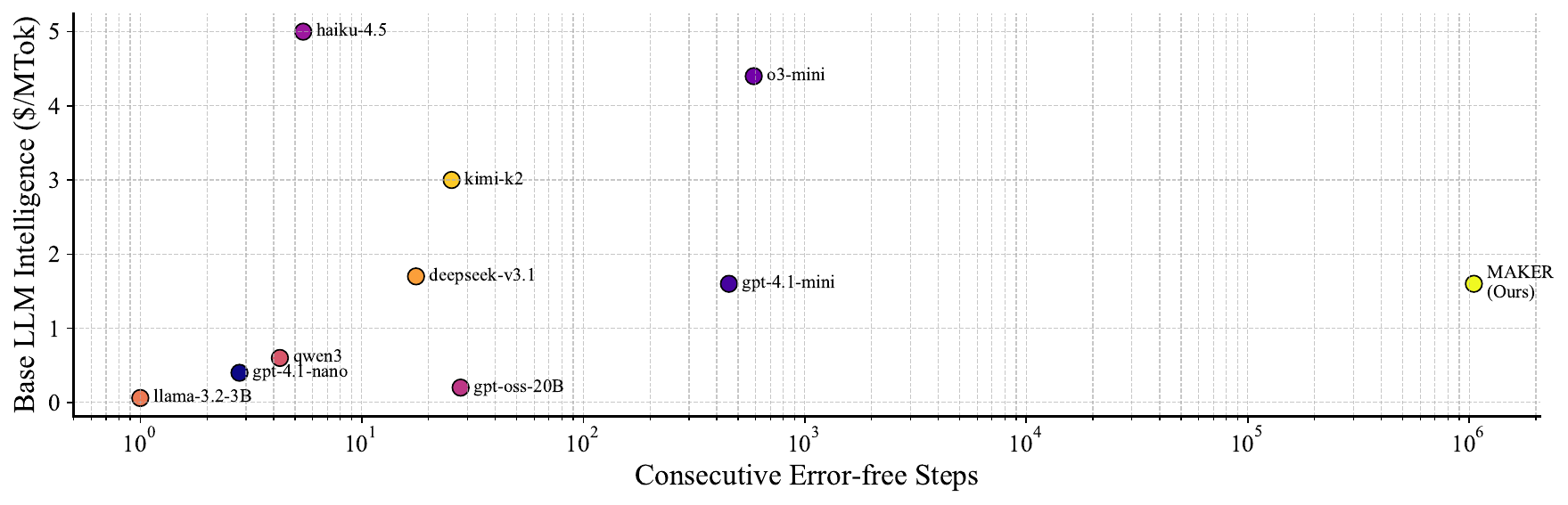}
    \captionof{figure}{
    This is the same figure as \ref{fig:intelligence_vs_scale}, but with a log-scale x-axis.
    }
    \label{fig:fig1_log_scale}

\section{Additional Derivations}
\label{app:derivations}
This section provides details for the derivation of Eq.~\ref{eq:k_min}.
Suppose the target probability for solving the full task with zero errors is $t$.
The goal is to find the minimum $k$ such that
\begin{equation}
    t \geq p_{\textrm{full}} = \Bigg(1 + \bigg(\frac{1-p}{p}\bigg)^k\Bigg)^{-\frac{s}{m}} \implies t^{-\frac{m}{s}} \geq 1 + \left(\frac{1-p}{p}\right)^k.
\end{equation}
Plugging in $a = t^{-\frac{m}{s}}$ and $b = \frac{1-p}{p}$ gives
\begin{equation}
    a \geq 1 + b^k \implies
    a -1 \geq b^k \implies
    \ln(a-1) \geq k \ln b \implies k \geq \frac{\ln(a-1)}{\ln b},
\end{equation}
since $\ln b = \ln\left(\frac{1-p}{p}\right) < 0$ when $p > 0.5$.
Replacing $a$ and $b$ and taking the first satisfying integer yields
\begin{equation}
    k_{\min} \;=\; \left\lceil \frac{\ln\left(t^{-m/s} - 1\right)}{\ln\left(\frac{1-p}{p}\right)} \right\rceil .
\end{equation}
To understand the asymptotic behavior of $k_{\min}$, first note
\begin{equation}
    t^{-\frac{m}{s}} = e^{-\frac{m}{s}\ln t} = e^x,
\end{equation}
where $x = -\frac{m}{s}\ln t$.
Suppose $t > e^{-1}$ (in our experiments it is close to 1).
Then, $x \in (0, 1)$, and the classic bounds hold:
\begin{align}
    x &\leq e^x - 1 \leq ex\\
    \implies -\frac{m}{s} \ln t &\leq t^{-\frac{m}{s}} - 1 \leq -\frac{em}{s} \ln t\\
    \implies \ln\left(-\frac{m}{s} \ln t\right) &\leq \ln\left(t^{-\frac{m}{s}} - 1\right) \leq \ln\left(-\frac{em}{s} \ln t\right)\\
    \implies \ln\left(-m \ln(t)\right) - \ln(s) &\leq \ln\left(t^{-\frac{m}{s}} - 1\right) \leq \ln\left(-em \ln(t)\right) - \ln(s)\\
    \implies \frac{\ln\left(-m \ln(t)\right) - \ln(s)}{\ln\left(\frac{1-p}{p}\right)} &\geq \frac{\ln\left(t^{-\frac{m}{s}} - 1\right)}{\ln\left(\frac{1-p}{p}\right)} \geq \frac{\ln\left(-em \ln(t)\right) - \ln(s)}{\ln\left(\frac{1-p}{p}\right)}\\
    \left\lceil \frac{\Theta(1) - \ln(s)}{-\Theta(1)} \right\rceil &\geq k_{\min} \geq \left\lceil \frac{\Theta(1) - \ln(s)}{-\Theta(1)} \right\rceil\\
    \implies k_{\min} &= \Theta(\ln(s)).
\end{align}

\section{Prompts and Parsers}
\label{app:prompts_and_parsers}

This section provides python code for the prompt templates ($\psi$) and parsers ($\psi_a$ and $\psi_x$) used in the experiments in Section~\ref{sec:results}.
The prompts are based on ones used in prior work \cite{shojaee2025illusion}.
\\

\emph{Prompt templates:}
\begin{CodeBox}
SYSTEM_PROMPT = """
You are a helpful assistant. Solve this puzzle for me.

There are three pegs and n disks of different sizes stacked on the first peg. The disks are
numbered from 1 (smallest) to n (largest). Disk moves in this puzzle should follow:
1. Only one disk can be moved at a time.
2. Each move consists of taking the upper disk from one stack and placing it on top of
another stack.
3. A larger disk may not be placed on top of a smaller disk.
The goal is to move the entire stack to the third peg.

Example: With 3 disks numbered 1 (smallest), 2, and 3 (largest), the initial state is [[3, 2, 1], [], []], and a solution might be:
moves = [[1, 0, 2], [2, 0, 1], [1, 2, 1], [3, 0, 2], [1, 1, 0], [2, 1, 2], [1, 0, 2]]
This means: Move disk 1 from peg 0 to peg 2, then move disk 2 from peg 0 to peg 1, and so on.

Requirements:
- The positions are 0-indexed (the leftmost peg is 0).
- Ensure your answer includes a single next move in this EXACT FORMAT:
```move = [disk id, from peg, to peg]```
- Ensure your answer includes the next state resulting from applying the move to the current state in this EXACT FORMAT:
```next_state = [[...], [...], [...]]```
"""

USER_TEMPLATE = """
Rules:
- Only one disk can be moved at a time.
- Only the top disk from any stack can be moved.
- A larger disk may not be placed on top of a smaller disk.

For all moves, follow the standard Tower of Hanoi procedure:
If the previous move did not move disk 1, move disk 1 clockwise one peg (0 -> 1 -> 2 -> 0).
If the previous move did move disk 1, make the only legal move that does not involve moving disk1.
Use these clear steps to find the next move given the previous move and current state.

Previous move: {previous_move}

Current State: {current_state}

Based on the previous move and current state, find the single next move that follows the procedure and the resulting next state.
"""
\end{CodeBox}

\emph{Repairing parser:}
\begin{CodeBox}
import re, ast

def extract_balanced_brackets(text, start_idx):
    """Extract a substring with balanced brackets [[...]] starting at start_idx"""
    bracket_stack = []
    i = start_idx
    while i < len(text):
        if text[i] == '[':
            bracket_stack.append('[')
        elif text[i] == ']':
            if not bracket_stack:
                break
            bracket_stack.pop()
            if not bracket_stack:
                return text[start_idx:i + 1]
        i += 1
    return text[start_idx:i] + ']'

def parse_move_state_repair(response_text):
    try:
        move_matches = re.findall(r"(?i)\bmove\b\s*=\s*(\[[^\[\]]*\])", response_text)
        if not move_matches:
            raise ValueError("No 'move' found in response.")
        move = ast.literal_eval(move_matches[-1].strip())
    except Exception as e:
        raise ValueError("Could not parse 'move' from response.") from e

    try:
        # Match last occurrence of 'next_state = [ [' with any whitespace
        pattern = re.compile(r"(?i)\bnext_state\b\s*=\s*(\[\s*\[)", re.DOTALL)
        matches = list(pattern.finditer(response_text))
        if not matches:
            raise ValueError("No 'next_state' found in response.")
        start_idx = matches[-1].start(1)  # last match
        next_state_str = extract_balanced_brackets(response_text, start_idx).strip()
        next_state = ast.literal_eval(next_state_str)
    except Exception as e:
        raise ValueError("Could not parse 'next_state' from response.") from e

    return move, next_state
\end{CodeBox}

\emph{Red-flagging parser:}
\begin{CodeBox}
import re, ast

def _validate_move(move):
    if not isinstance(move, list) or len(move) != 3 or not all(isinstance(x, int) for x in move):
        raise ValueError("'move' must be a list of exactly 3 integers.")
    return move

def _validate_state(state):
    if not (isinstance(state, list) and len(state) == 3 and all(isinstance(t, list) for t in state)):
        raise ValueError("'next_state' must be a list of three lists.")
    flat = [x for t in state for x in t]
    if not all(isinstance(x, int) for x in flat):
        raise ValueError("All entries in 'next_state' must be integers.")
    if len(flat) != 20 or set(flat) != set(range(1, 21)):
        missing = sorted(set(range(1, 21)) - set(flat))
        extra   = sorted(set(flat) - set(range(1, 21)))
        raise ValueError("State must contain 1..20 exactly once. "
                         f"Missing: {missing or '[]'}, Extras: {extra or '[]'}")
    return state

def parse_move_state_flag(response_text: str):
    # Match square brackets
    move_pat = re.compile(r"(?is)\bmove\b\s*=\s*(\[[^\[\]]*\])")
    state_pat = re.compile(
        r"(?is)\bnext_state\b\s*=\s*(\[\s*\[[^\[\]]*\]\s*,\s*\[[^\[\]]*\]\s*,\s*\[[^\[\]]*\]\s*\])"
    )

    move_matches = list(move_pat.finditer(response_text))
    if not move_matches:
        raise ValueError("No 'move = [...]' found.")
    move_str = move_matches[-1].group(1)  # last 'move'

    state_matches = list(state_pat.finditer(response_text))
    if not state_matches:
        raise ValueError("No 'next_state = [[...],[...],[...]]' found.")
    state_str = state_matches[-1].group(1)  # last 'next_state'

    try:
        move = ast.literal_eval(move_str)
    except Exception as e:
        raise ValueError("Could not parse 'move' as a Python list.") from e
    try:
        next_state = ast.literal_eval(state_str)
    except Exception as e:
        raise ValueError("Could not parse 'next_state' as Python lists.") from e

    return _validate_move(move), _validate_state(next_state)


\end{CodeBox}
\section{Sample Responses}
\label{app:sample_response}

\begin{figure}[h]
    \centering
    \includegraphics[width=0.7\linewidth]{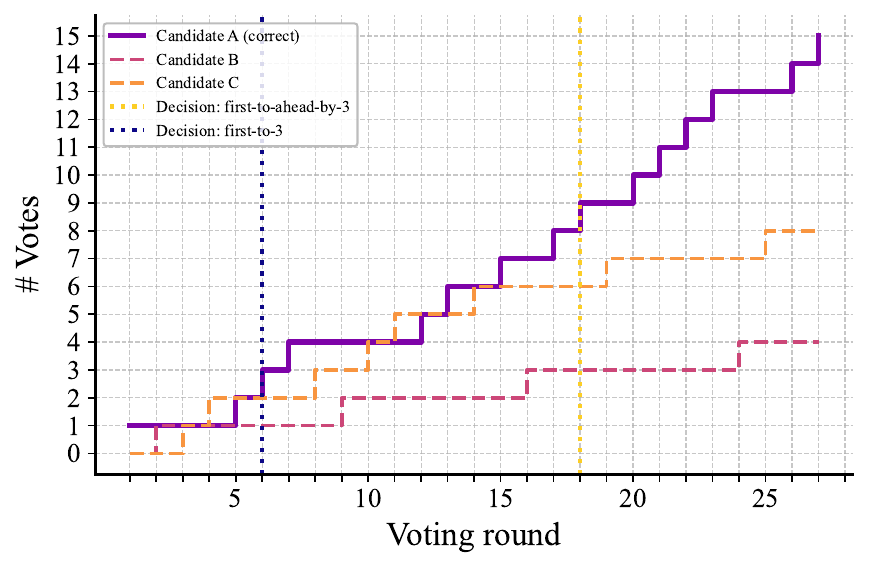}
    \caption{
    \emph{Vote race for step 10241.}
    This figure depicts the vote race for the pathological step 102421 that requires far more votes than any other step, though the correct decision is eventually made under both voting rules (first-to-$k$ and first-to-ahead-by-$k$; Figure~\ref{fig:full_solution}).
    Sample responses leading to each candidate are shown above.
    Additional samples were drawn after the decision was made to confirm that Candidate A does keep pulling further ahead.
    Although the correct decision was made, the fact that this pathological sample exists serves as motivation for developing more sophisticated error decorrelation methods in the future (Section~\ref{sec:discussion}).
    }
    \label{fig:pathology}
\end{figure}

This section provides sample responses from the full experiment in Section~\ref{sec:solving_the_full_problem}.
To give a sense of the kind of behavior that can occur, a shortest and a longest sample (with respect to number of tokens) from the first round of voting are shown, along with samples for the three racing candidates in the pathological step, step 10241, that takes 18 votes.
Figure~\ref{fig:pathology} illustrates the process of that race.\\

\emph{Short sample, 256 Tokens, Step 950202:}\\

This sample demonstrates straightforward reasoning.

\begin{CodeBox}
Let's analyze the situation step-by-step:

Previous move: [2, 2, 1] (disk 2 moved)

Since the previous move did move disk 2 (not disk 1), the next move should be to move disk 1 clockwise one peg.

Disk 1 is currently on peg 0 (top disk of peg 0 is 1).

Clockwise from peg 0 is peg 1.

So, move disk 1 from peg 0 to peg 1.

Now, let's update the state:

- Remove disk 1 from peg 0: peg 0 becomes [6, 5, 4]
- Add disk 1 to peg 1 on top: peg 1 becomes [17, 16, 7, 2, 1]
- Peg 2 remains the same.

So the next move and next state are:

``` 
move = [1, 0, 1]
next_state = [[6, 5, 4], [17, 16, 7, 2, 1], [20, 19, 18, 15, 14, 13, 12, 11, 10, 9, 8, 3]]
```
\end{CodeBox}

\emph{Long sample, 2048 Tokens, Step 539011:}\\

This sample demonstrates confusion.
After it makes an error in its early reasoning, it talks in circles (``Wait, maybe the stacks are not as we think.'') until it hits the max token limit on the API call before outputting a correctly-formatted answer.

\begin{CodeBox}
Let's analyze the situation step-by-step:

- Previous move was disk 1 from peg 0 to peg 1.
- According to the rules:
  - If the previous move did move disk 1, the next move should be the only legal move that does not involve moving disk 1.

Current state:
- Peg 0: [14, 13, 12, 9, 8]
- Peg 1: [19, 18, 17, 16, 15, 2, 1]
- Peg 2: [20, 11, 10, 7, 6, 5, 4, 3]

Top disks on each peg:
- Peg 0 top disk: 8
- Peg 1 top disk: 1 (disk 1 is on top here)
- Peg 2 top disk: 3

We need to find a legal move that does not involve disk 1.

Possible moves (excluding disk 1):

- Move top disk from peg 0 (disk 8) to peg 1:
  - Peg 1 top disk is 1, which is smaller than 8, so cannot place 8 on 1.
- Move top disk from peg 0 (disk 8) to peg 2:
  - Peg 2 top disk is 3, which is smaller than 8, so cannot place 8 on 3.
- Move top disk from peg 1 (disk 2) to peg 0:
  - Peg 0 top disk is 8, which is larger than 2, so can place 2 on 8.
- Move top disk from peg 1 (disk 2) to peg 2:
  - Peg 2 top disk is 3, which is larger than 2, so can place 2 on 3.
- Move top disk from peg 2 (disk 3) to peg 0:
  - Peg 0 top disk is 8, which is larger than 3, so can place 3 on 8.
- Move top disk from peg 2 (disk 3) to peg 1:
  - Peg 1 top disk is 2, which is smaller than 3, so cannot place 3 on 2.

Legal moves excluding disk 1 are:
- Move disk 2 from peg 1 to peg 0
- Move disk 2 from peg 1 to peg 2
- Move disk 3 from peg 2 to peg 0

Among these, the only legal move that does not involve disk 1 and is the only legal move is the one that does not involve disk 1 and is unique.

We have three legal moves, so we need to check which one is the only legal move that does not involve disk 1.

Wait, the instruction says: "If the previous move did move disk 1, make the only legal move that does not involve moving disk 1."

So, if there are multiple legal moves excluding disk 1, we need to find the only legal move.

But here, there are multiple legal moves excluding disk 1.

Is there a mistake?

Let's check the rules again.

The standard Tower of Hanoi procedure for 3 pegs and n disks is:

- If the previous move did not move disk 1, move disk 1 clockwise one peg.
- If the previous move did move disk 1, make the only legal move that does not involve moving disk 1.

So, the "only legal move" means there should be exactly one legal move excluding disk 1.

But here, we have three legal moves excluding disk 1.

Is there a mistake in the analysis?

Let's check the top disks again carefully.

Peg 0: top disk 8

Peg 1: top disk 1

Peg 2: top disk 3

Possible moves excluding disk 1:

- From peg 0 (disk 8) to peg 1 (top disk 1): cannot place 8 on 1 (larger on smaller) - illegal

- From peg 0 (disk 8) to peg 2 (top disk 3): cannot place 8 on 3 - illegal

- From peg 1 (disk 2) to peg 0 (top disk 8): 2 < 8, legal

- From peg 1 (disk 2) to peg 2 (top disk 3): 2 < 3, legal

- From peg 2 (disk 3) to peg 0 (top disk 8): 3 < 8, legal

- From peg 2 (disk 3) to peg 1 (top disk 2): 3 > 2, illegal

So, three legal moves excluding disk 1.

But the instruction says "make the only legal move that does not involve moving disk 1."

So, the only legal move is ambiguous here.

Is there a standard order to pick the move?

In the Tower of Hanoi standard procedure, the non-disk 1 move is unique because only one move is legal at that point.

But here, with many disks, the situation is more complex.

Given the standard Tower of Hanoi procedure, the non-disk 1 move is the only legal move that does not involve disk 1.

So, we need to find the only legal move excluding disk 1.

Since there are multiple, maybe the standard procedure is to pick the move involving the smallest disk other than disk 1.

Among disks 2 and 3, disk 2 is smaller.

Between moving disk 2 from peg 1 to peg 0 or peg 2, or moving disk 3 from peg 2 to peg 0.

Is moving disk 2 from peg 1 to peg 0 legal? Yes.

Is moving disk 2 from peg 1 to peg 2 legal? Yes.

Is moving disk 3 from peg 2 to peg 0 legal? Yes.

But moving disk 2 to peg 0 or peg 2, which is the correct move?

In the standard Tower of Hanoi, the non-disk 1 move is the only legal move that does not involve disk 1.

So, maybe the only legal move is the one that does not involve disk 1 and is the only legal move at that time.

So, maybe the other moves are illegal because they would violate the rules.

Wait, maybe the stacks are not as we think.

Are the disks ordered from bottom to top or top to bottom?

The initial state is given as [[3, 2, 1], [], []], which means the bottom disk is first in the list, top disk last.

So, the last element in the list is the top disk.

So, in the current state:

Peg 0: [14, 13, 12, 9, 8] -> top disk is 8

Peg 1: [19, 18, 17, 16, 15, 2, 1] -> top disk is 1

Peg 2: [20, 11, 10, 7, 6, 5, 4, 3] -> top disk is 3

So, the top disks are as above.

Now, let's check the moves again.

- Move disk 8 from peg 0 to peg 1: peg 1 top disk is 1, 8 > 1, illegal.

- Move disk 8 from peg 0 to peg 2: peg 2 top disk is 3, 8 > 3, illegal.

- Move disk 2 from peg 1 to peg 0: peg 0 top disk is 8, 2 < 8, legal.

- Move disk 2 from peg 1 to peg 2: peg 2 top disk is 3, 2 < 3, legal.

- Move disk 3 from peg 2 to peg 0: peg 0 top disk is 8, 3 < 8, legal.

- Move disk 3 from peg 2 to peg 1: peg 1 top disk is 2, 3 > 2, illegal.

So, three legal moves excluding disk 1.

But the instruction says "make the only legal move that does not involve moving disk 1."

So, maybe the standard Tower of Hanoi procedure is that the non-disk 1 move is unique.

Given that, maybe the only legal move is the one that moves the smallest disk other than disk 1.

Disk 2 is smaller than disk 3.

Between moving disk 2 from peg 1 to peg 0 or peg 2, which is the correct move?

In the standard Tower of Hanoi, the non-disk 1 move is the only legal move that does not involve disk 1.

So, maybe the only legal move is moving disk 2 from peg 1 to peg 0.

Let's pick that move.

Move = [2, 1, 0]

Now, apply this move to the current state:

- Remove disk 2 from peg 1: peg 1 becomes [19, 18, 17, 16, 15, 1]

- Add disk 2 to peg 0: peg 0 becomes [14, 13, 12, 9, 8, 2]

Peg 2 remains the same.

So, next_state = [[14, 13, 12, 9, 8, 2], [19, 18, 17, 16, 15, 1], [20, 11, 10, 7, 6, 5, 4, 3]]

Final answer:

``` 
move = [2, 1, 0]
next_state = [[14, 13, 12, 9, 
\end{CodeBox}

\emph{Step 10241, Candidate A (correct) sample:}
\begin{CodeBox}
Let's analyze the problem step-by-step:

Previous move: [1, 2, 0] (disk 1 moved from peg 2 to peg 0)

Current state:
peg 0: [20, 19, 18, 17, 16, 15, 12, 1]
peg 1: [13]
peg 2: [14, 11, 10, 9, 8, 7, 6, 5, 4, 3, 2]

According to the rules:

- If the previous move did move disk 1, make the only legal move that does not involve moving disk 1.

Since the previous move did move disk 1, we must make the only legal move that does not involve moving disk 1.

Let's find the top disks on each peg:

- peg 0 top disk: 1 (smallest)
- peg 1 top disk: 13
- peg 2 top disk: 2

We cannot move disk 1 this time.

Possible moves not involving disk 1:

- Move disk 13 from peg 1 to peg 0 or peg 2
- Move disk 2 from peg 2 to peg 0 or peg 1

Check legality:

- Move disk 13 from peg 1 to peg 0: peg 0 top disk is 1, which is smaller than 13, so illegal (larger disk cannot be placed on smaller disk).
- Move disk 13 from peg 1 to peg 2: peg 2 top disk is 2, which is smaller than 13, so illegal.
- Move disk 2 from peg 2 to peg 0: peg 0 top disk is 1, which is smaller than 2, so illegal.
- Move disk 2 from peg 2 to peg 1: peg 1 top disk is 13, which is larger than 2, so legal.

Therefore, the only legal move not involving disk 1 is moving disk 2 from peg 2 to peg 1.

Apply this move:

Move = [2, 2, 1]

Update state:

- Remove disk 2 from peg 2: peg 2 becomes [14, 11, 10, 9, 8, 7, 6, 5, 4, 3]
- Add disk 2 on top of peg 1: peg 1 becomes [13, 2]

Final next state:

peg 0: [20, 19, 18, 17, 16, 15, 12, 1]

peg 1: [13, 2]

peg 2: [14, 11, 10, 9, 8, 7, 6, 5, 4, 3]

Answer:

``` 
move = [2, 2, 1]
next_state = [[20, 19, 18, 17, 16, 15, 12, 1], [13, 2], [14, 11, 10, 9, 8, 7, 6, 5, 4, 3]]
```
\end{CodeBox}

\emph{Step 10241, Candidate B (incorrect) sample:}
\begin{CodeBox}
Let's analyze the situation step-by-step:

Previous move: [1, 2, 0] (disk 1 was moved from peg 2 to peg 0)

According to the rules:
- If the previous move did move disk 1, make the only legal move that does not involve moving disk 1.

Current state:
Peg 0: [20, 19, 18, 17, 16, 15, 12, 1]
Peg 1: [13]
Peg 2: [14, 11, 10, 9, 8, 7, 6, 5, 4, 3, 2]

Disk 1 is on peg 0, top disk.

We need to find the only legal move that does not involve moving disk 1.

Check possible moves between pegs 0, 1, and 2, excluding disk 1:

- From peg 0 (top disk is 1) to peg 1 (top disk is 13): can't move disk 1, so no move from peg 0.
- From peg 1 (top disk is 13) to peg 0 (top disk is 1): disk 13 cannot be placed on disk 1 (smaller), so no.
- From peg 1 (top disk is 13) to peg 2 (top disk is 14): disk 13 can be placed on disk 14 (larger), legal move.
- From peg 2 (top disk is 14) to peg 0 (top disk is 1): disk 14 cannot be placed on disk 1 (smaller), no.
- From peg 2 (top disk is 14) to peg 1 (top disk is 13): disk 14 cannot be placed on disk 13 (smaller), no.

So the only legal move not involving disk 1 is moving disk 13 from peg 1 to peg 2.

Perform this move:

Move = [13, 1, 2]

Update state:

Peg 0: [20, 19, 18, 17, 16, 15, 12, 1] (unchanged)
Peg 1: [] (removed 13)
Peg 2: [13, 14, 11, 10, 9, 8, 7, 6, 5, 4, 3, 2] (13 placed on top)

So the next state is:

[[20, 19, 18, 17, 16, 15, 12, 1], [], [13, 14, 11, 10, 9, 8, 7, 6, 5, 4, 3, 2]]

Final answer:

``` 
move = [13, 1, 2]
next_state = [[20, 19, 18, 17, 16, 15, 12, 1], [], [13, 14, 11, 10, 9, 8, 7, 6, 5, 4, 3, 2]]
```
\end{CodeBox}

\emph{Step 10241, Candidate C (incorrect) sample:}
\begin{CodeBox}
Let's analyze the problem step-by-step:

- Previous move: [1, 2, 0] (disk 1 was moved)
- Current state:
  Peg 0: [20, 19, 18, 17, 16, 15, 12, 1]
  Peg 1: [13]
  Peg 2: [14, 11, 10, 9, 8, 7, 6, 5, 4, 3, 2]

According to the rules:

1. If the previous move did move disk 1, then the next move must be the only legal move that does not involve moving disk 1.

So, we need to find the only legal move that does not involve disk 1.

Disk 1 is on peg 0 (top disk on peg 0 is disk 1).

We cannot move disk 1 this turn.

We need to find a legal move involving disks other than disk 1.

Let's check the top disks on each peg:

- Peg 0: top disk is 1 (cannot move disk 1)
- Peg 1: top disk is 13
- Peg 2: top disk is 14

Possible moves (excluding disk 1):

- Move disk 13 from peg 1 to peg 0 or peg 2
- Move disk 14 from peg 2 to peg 0 or peg 1

Check legality:

- Moving disk 13 to peg 0: top disk on peg 0 is 1 (smaller), cannot place larger disk on smaller disk.
- Moving disk 13 to peg 2: top disk on peg 2 is 14 (larger), so disk 13 can be placed on 14.
- Moving disk 14 to peg 0: top disk on peg 0 is 1 (smaller), cannot place larger disk on smaller disk.
- Moving disk 14 to peg 1: top disk on peg 1 is 13 (smaller), cannot place larger disk on smaller disk.

So the only legal move is moving disk 13 from peg 1 to peg 2.

Perform this move:

- Remove disk 13 from peg 1
- Add disk 13 on top of peg 2

New state:

Peg 0: [20, 19, 18, 17, 16, 15, 12, 1]

Peg 1: []

Peg 2: [14, 11, 10, 9, 8, 7, 6, 5, 4, 3, 2, 13]

Answer:

``` 
move = [13, 1, 2]
next_state = [[20, 19, 18, 17, 16, 15, 12, 1], [], [14, 11, 10, 9, 8, 7, 6, 5, 4, 3, 2, 13]]
```
\end{CodeBox}

\section{Open-source Model Details}
\label{app:model_endpoints}

The below table gives details on the open-source models used in this paper, which were accessed via the \url{together.ai} API.
Temperature 0.1 was used for all open-source models.

\begin{table}[h!]
\centering
\begin{tabular}{lrlrr}
\toprule
\textbf{Model Name} & \textbf{\# Params} & \textbf{Endpoint} & \textbf{Input \$/MTok} & \textbf{Output \$/MTok} \\
\midrule
Qwen-3 & 235B & Qwen/Qwen3-235B-A22B-Instruct-2507-tput & 0.2 & 0.6 \\
DeepSeek-v3.1 & 671B & deepseek-ai/DeepSeek-V3 & 0.6 & 1.7 \\
Kimi-K2 & 1T & moonshotai/Kimi-K2-Instruct & 1.0 & 3.0 \\
GPT-OSS-20B & 20B & OpenAI/gpt-oss-20B & 0.05 & 0.2 \\
Llama-3.2-3B & 3.2B & meta-llama/Llama-3.2-3B-Instruct-Turbo & 0.06 & 0.06 \\
\bottomrule
\end{tabular}
\vspace{5pt}
\caption{\emph{Open-source model details.} Models accessed through \url{together.ai} API.}
\label{tab:model_pricing}
\end{table}

\section{Multiplication Experiments}
\label{app:accuracy_vs_k}
\includegraphics[width=\linewidth]{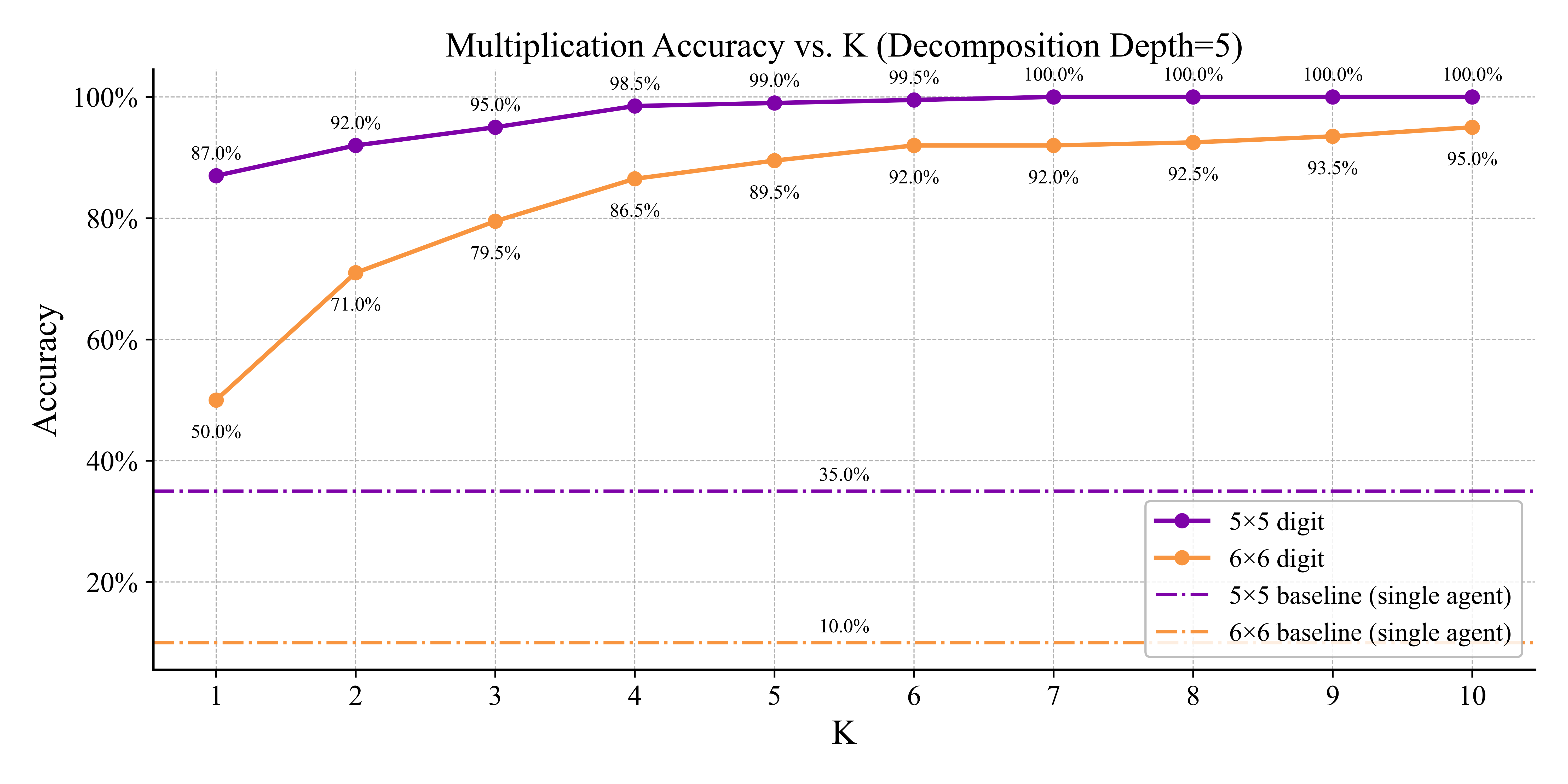}
\captionof{figure}{
Solve rate of the multi-agent system on 5×5 and 6×6 digit multiplication tasks as a function of voting parameter $k$, with decomposition depth fixed at $5$. Dotted horizontal lines indicate baseline single-agent performance for each task. As $k$ increases, accuracy improves for both 5×5 and 6×6 multiplication, reaching a perfect solve rate for 5x5 and reaching the target solve rate used in Section~\ref{sec:results} of $t = 0.95$ for 6x6.
These results demonstrate the benefit of voting even at fixed reasoning depth (Algorithm~\ref{alg:recursive-multiagent}).
In this experiment, \texttt{gpt-4.1-mini} was used for all agents.\\
}
\label{fig:mul_accuracy_vs_k}

Bai \emph{et al.}~\cite{bai2025canttransformerslearnmultiplication} showed that standard Transformers struggle with multi-digit multiplication because attention alone fails to maintain long-range dependencies between intermediate digit interactions. Their reverse-engineering analysis revealed that successful computation requires constructing a directed acyclic “attention tree” to propagate partial products across steps. The MDAP implementation, validated here on the same multiplication benchmark, is more general: it recursively decomposes any given task into subtasks and mitigates the multi-step degradation of accuracy by voting on each decomposition step, composition step, and atomic reasoning step. In contrast to model-specific architectural fixes, this voting-based recursive reasoning mechanism scales naturally with task complexity, allowing reliable multi-step inference beyond the arithmetic domain (Algorithm~\ref{alg:recursive-multiagent}). The source code for this experiment is available here: \url{www.github.com/cognizant-ai-lab/neuro-san-benchmarking}.

\begin{algorithm}[H]
\caption{Recursive multi-agent solve: decomposition sampling + voting until non-decomposable or depth limit, then solution sampling + voting, recursively composed to a final answer}
\label{alg:recursive-multiagent}
\begin{algorithmic}[1]
\State $N \gets 2k-1$ \Comment{First-to-$k$ voting, $N$ candidates per step}
\Function{Decompose}{$x$} \State sample $N$ decompositions via \textsc{Decomposer}$(x)$; vote via \textsc{SolutionDiscriminator} until one reaches $k$ (else argmax); return $(P_1,P_2,C)$ \EndFunction
\Function{Atomic}{$x$} \State sample $N$ answers via \textsc{ThinkingModule}$(x)$; vote via \textsc{CompositionDiscriminator}; return winner \EndFunction
\Function{Solve}{$x,d$}
  \If{$d \ge \text{MAX\_DEPTH}$} \State \Return \Call{Atomic}{$x$} \EndIf
  \State $(P_1,P_2,C) \gets \Call{Decompose}{x}$
  \If{$P_1=\varnothing$ or $P_2=\varnothing$ or $C=\varnothing$} \State \Return \Call{Atomic}{$x$} \EndIf
  \State $s_1 \gets \Call{Solve}{P_1,d+1}$,\quad $s_2 \gets \Call{Solve}{P_2,d+1}$
  \State sample $N$ composed solutions via \textsc{ThinkingModule}$(\text{``Solve }C(P_1,P_2)\text{ with }P_1{=}s_1,P_2{=}s_2\text{''})$
  \State vote via \textsc{CompositionDiscriminator} until one reaches $k$ (else argmax); \Return winner
\EndFunction
\end{algorithmic}
\end{algorithm}

\end{document}